%% file: iclr2025_conference.tex
\documentclass{article} 
\usepackage{iclr2025_conference,times}

\input{math_commands.tex}

\usepackage{hyperref}
\usepackage{url}

\usepackage{graphicx}
\usepackage{algorithm}
\usepackage{algorithmic}
\usepackage{amsmath}

\usepackage{subfigure}
\usepackage{booktabs}
\usepackage{multirow}
\usepackage{color}

\usepackage{wrapfig}
\usepackage{caption}
\usepackage{fontawesome5}
\usepackage{amssymb}
\usepackage{xcolor}
\usepackage{paralist}
\usepackage{colortbl}

\definecolor{YellowGreen}{rgb}{0.6, 0.8, 0.2}

\newenvironment{itemize*}%
 {\leftmargini=10pt\begin{compactitem}%
  \setlength{\itemsep}{0pt}%
  \setlength{\parskip}{0pt}%
  }%
 {\end{compactitem}}
\newenvironment{enumerate*}%
 {\begin{enumerate}%
  \setlength{\itemsep}{0pt}%
  \setlength{\parskip}{0pt}}%
 {\end{enumerate}}

\title{\textit{CHiP}: Cross-modal Hierarchical Direct Preference Optimization for Multimodal LLMs}

\author{
Jinlan Fu$^{1}$  \hspace{.7em}
Shenzhen Huangfu$^{1,2}$\thanks{Work done during an internship at NUS.}   \hspace{.7em}
Hao Fei$^{1}$ \hspace{.7em} 
Xiaoyu Shen$^{3}$ \hspace{.7em} \\
\textbf{
Bryan Hooi$^{1}$ \hspace{.7em} 
Xipeng Qiu$^{2}$\thanks{Corresponding author.} \hspace{.7em} 
See-Kiong Ng$^{1}$ \hspace{.7em} 
}
\\
[1ex]
$^{1}$National University of Singapore \quad $^{2}$Fudan University \\
$^{3}$Digital Twin Institute, Eastern Institute of Technology, Ningbo\\
[1ex]
\small{
\texttt{\,jinlanjonna@gmail.com, shenzhenhuangfu@gmail.com}
}
}

\iclrfinalcopy 
\begin{document}

\maketitle

\begin{abstract}
Multimodal Large Language Models (MLLMs) still struggle with hallucinations despite their impressive capabilities. Recent studies have attempted to mitigate this by applying Direct Preference Optimization (DPO) to multimodal scenarios using preference pairs from text-based responses.
However, our analysis of representation distributions reveals that multimodal DPO struggles to align image and text representations and to distinguish between hallucinated and non-hallucinated descriptions. To address these challenges,
In this work, we propose a \underline{\textbf{C}}ross-modal \underline{\textbf{Hi}}erarchical Direct \underline{\textbf{P}}reference Optimization (CHiP) to address these limitations.
We introduce a visual preference optimization module within the DPO framework, enabling MLLMs to learn from both textual and visual preferences simultaneously. Furthermore, we propose a hierarchical textual preference optimization module that allows the model to capture preferences at multiple granular levels, including response, segment, and token levels.
We evaluate CHiP through both quantitative and qualitative analyses, with results across multiple benchmarks demonstrating its effectiveness in reducing hallucinations. On the Object HalBench dataset, CHiP outperforms DPO in hallucination reduction, achieving improvements of 52.7\% and 55.5\% relative points based on the base model Muffin and LLaVA models, respectively. We make all our datasets and code publicly available. 
\footnote{\url{https://github.com/LVUGAI/CHiP}}

\end{abstract}

\section{Introduction}

The emergence of large language models (LLMs) has demonstrated unprecedented intelligence  \citep{chiang2023vicuna,touvron2023llama,llama3modelcard}, bringing us closer to achieving human-level AI. 
Concurrently, building on the foundation of text-based LLMs, research in Multimodal Large Language Models (MLLMs) has also rapidly surged, leading to the development of powerful multimodal models such as GPT-4V \citep{gpt4V}, BLIP2 \citep{0008LSH23}, and LLaVA \citep{liu2024visual}. 
The current MLLMs typically integrate the visual encoder into the text-oriented backbone LLMs through a connector to achieve the understanding of visual signals~\citep{liu2024visual,yao2024minicpmv}. 
Although MLLMs have achieved impressive results, hallucination remains a significant challenge, where the model's output is not based on the visual input~\citep{bai2024hallucination,jiang2024hallucination}.

\begin{figure}[t]
	\centering
    \subfigure[LLaVA]{
	\begin{minipage}[b]{0.2\textwidth}
	\includegraphics[width=1\textwidth]{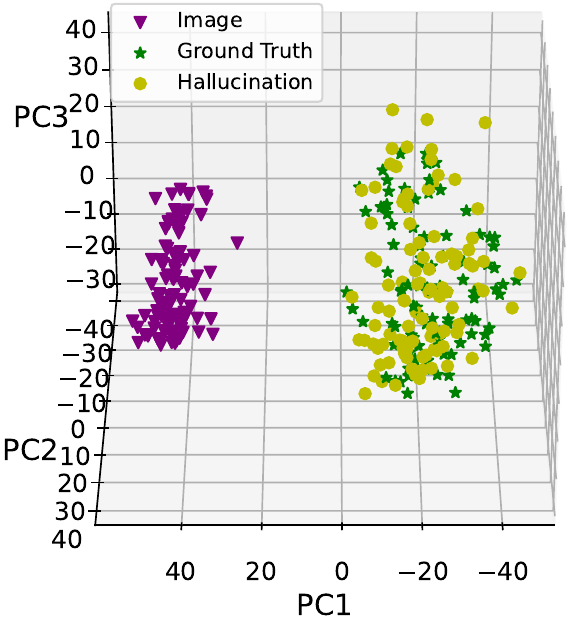}
	\end{minipage}
	}  
	\subfigure[LLaVA+DPO]{
	\begin{minipage}[b]{0.2 \textwidth}
		\includegraphics[width=1\textwidth]{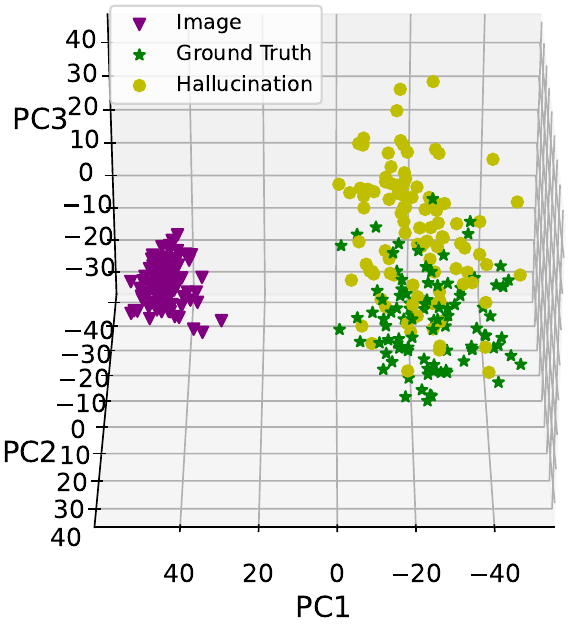}
	\end{minipage}
	} 
	\subfigure[LLaVA+CHiP]{
	\begin{minipage}[b]{0.2 \textwidth}
		\includegraphics[width=1\textwidth]{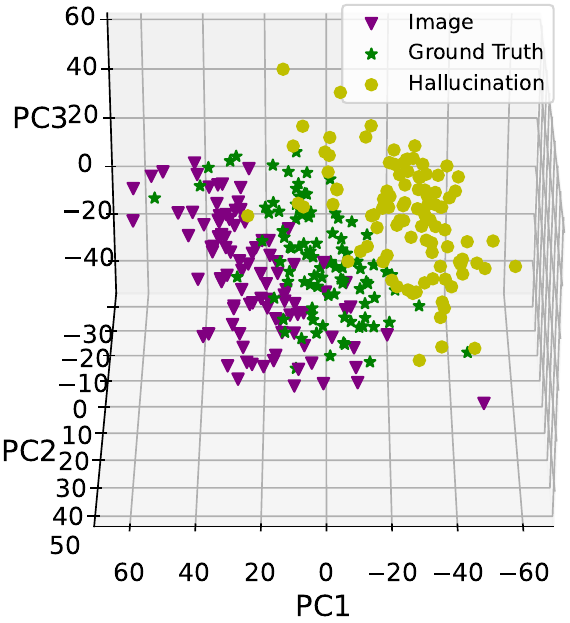}
	\end{minipage}
	} 
    \subfigure[Performance]{
	\begin{minipage}[b]{0.25 \textwidth}
		\includegraphics[width=1\textwidth]{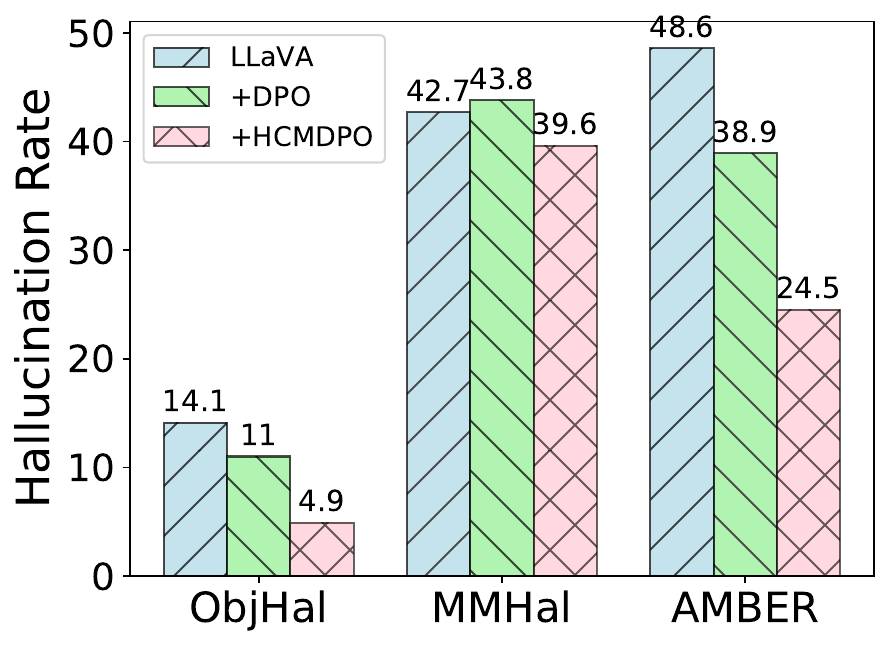}
	\end{minipage}
	} 
 \vspace{-6pt}
	\caption{Comparison of representation distributions and performance across models. Representations are constructed by selecting 150 samples (images, non-hallucinated descriptions, and hallucinated descriptions). The image or text semantics are represented using the last token embedding from the LLM.
     (d) is the hallucination rate (lower the better) comparison of different models on hallucination benchmarks, namely ObjHal, MMHal, and AMBER. Findings: 
 \textit{(1) DPO struggles to align image and description representations and to effectively distinguish between hallucinated and non-hallucinated descriptions.
(2) The proposed CHiP method, which incorporates both image and fine-grained text preferences, achieves better alignment between images and ground-truth descriptions while increasing the distance between ground-truth and hallucinated descriptions.
(3) CHiP outperforms DPO and original LLaVA in terms of hallucination rate.
}
}
\vspace{-4mm}
\label{fig:distribution}
\end{figure}

\begin{figure*}
    \centering
    \includegraphics[width=0.99\linewidth]{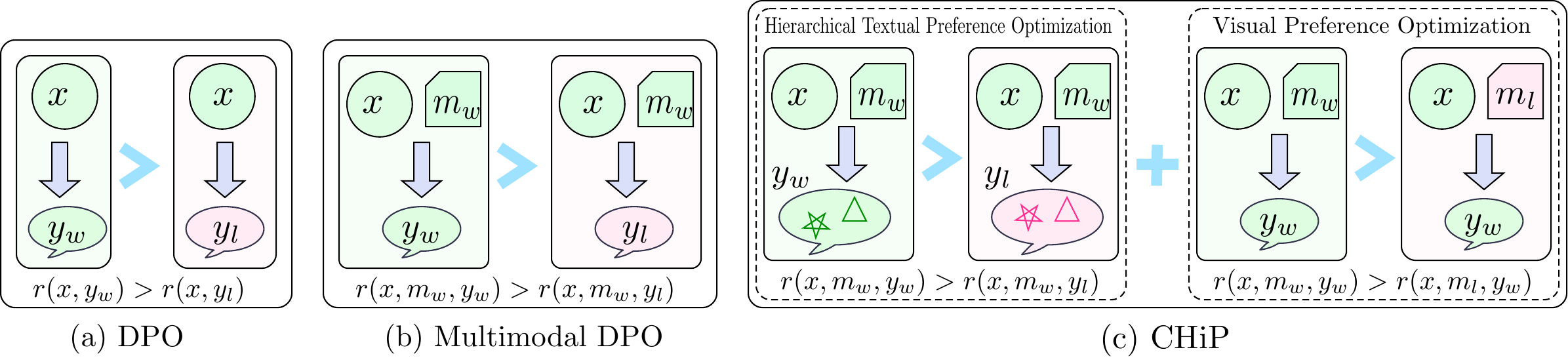}
    \vspace{-6pt}
    \caption{Comparison of preference optimization in different scenarios: (a) DPO, (b) Multimodal DPO, and (c) Cross-modal Hierarchical Direct Preference Optimization (CHiP).
    $x$ represents the instruction. 
    $y_w$ denotes the response preferred by the human over $y_l$.
    $m_w$ represents the image that is more likely to generate the preferred response $y_w$ than $m_l$. 
    \textcolor{YellowGreen}{$\bigstar$} (\textcolor{YellowGreen}{$\blacktriangle$}) and \textcolor{pink}{$\bigstar$} (\textcolor{pink}{$\blacktriangle$}) 
  represent the segments (tokens) involved in the hierarchy reward calculation in the preferred and unpreferred responses.}
    \label{fig:framework}
\vspace{-5mm}
\end{figure*}

With the help of Direct Preference Optimization (DPO)~\citep{rafailov2024direct} (\autoref{fig:framework}-(a)), text-oriented LLMs have achieved satisfactory alignment with human preferences, which can help prevent hallucinations and enhance their ability to meet human needs.
However, the alignment techniques for Multimodal LLMs (MLLMs) remain underexplored.
A natural approach is to extend DPO from the text modality to multimodal contexts through multimodal DPO~\citep{abs-2403-08730,abs-2405-18654} (\autoref{fig:framework}-(b)). 
However, simply replacing text preference data with multimodal preference data is insufficient to handle complex multimodal scenarios. 
In this work, we identify the limitations of multimodal DPO by visualizing the representation distributions of both images and texts and propose solutions to overcome these challenges.
Ideally, for well-aligned MLLMs, the representations of an image and its ground-truth description should be as close as possible, while the representations of ground-truth and hallucinated descriptions should be more distant. \autoref{fig:distribution} shows a visualization of the last token representations of image and text in the LLM (specifically LLaMA~\citep{touvron2023llama}) for LLaVA-1.6~\citep{liu2024llavanext}, using 150 samples (image, ground-truth description, and hallucinated description). 
\autoref{fig:distribution}-(a, b, d) highlights the limitations of multimodal DPO, showing its difficulty in aligning image and description representations and distinguishing between hallucinated and non-hallucinated descriptions, hindering performance improvement in hallucination evaluations.

To address these limitations, we propose \underline{C}ross-modal \underline{Hi}erarchical Direct \underline{P}reference Optimization (CHiP) 
 (\autoref{fig:framework}-(c)), which enhances the alignment from multiple textual granularities (e.g., response, segment, token levels) and visual preferences.
Specifically, 
we introduce \textit{Visual Preference Optimization} by constructing visual preference pairs, allowing the model to learn preferences from both text and visual modalities, aligning text and image representations more closely.
Moreover, we introduce a \textit{Hierarchical Textual Preference Optimization} to allow MLLMs to acquire preference information at multiple granular levels, namely, response, segment, and token, enhancing their ability to differentiate between hallucinated and non-hallucinated text.
To validate the efficacy of CHiP, we evaluate it on four popular hallucination benchmarks under LLaVA-1.6 and Muffin frameworks. The results show that CHiP outperforms GPT-4V significantly on the evaluated benchmarks.
Moreover, on the Object HalBench dataset, based on Muffin and LLaVA-1.6 models, CHiP outperforms DPO in hallucination reduction, with performance improvements of 52.7\% and 55.5\% relative points, respectively.

To sum up, our contributions are threefold:
\begin{itemize*}
    \item We analyze the limitations of multimodal DPO through image and text representation distributions, emphasizing its failure to achieve cross-modal  semantic alignment and distinguish between hallucinated and non-hallucinated descriptions.
    \item 
   We propose CHiP to address these limitations. CHiP includes a hierarchical textual preference optimization module to capture fine-grained  (i.e., response, segment, and token) preferences and a visual preference optimization module to extract cross-modal preferences.
    \item We equipped CHiP with various MLLMs, and the results of multiple datasets demonstrate that CHiP reduces hallucinations and enhances cross-modal semantic alignment.
\end{itemize*}

\section{Related Work}

Multimodal Large Language Models (MLLMs)~\citep{liu2024visual,Qwen-VL,InstructBLIP,he2024efficient} play a crucial role in multimodal understanding and reasoning tasks by processing both images and text. Their development has been fueled by progress in open-source LLMs~\citep{touvron2023llama,llama3modelcard,chiang2023vicuna} and cutting-edge image encoders~\citep{radford2021learning,wang2023image,li2022grounded}.

However, misalignment between images and text causes MLLMs to face issues like hallucinations and errors. Mitigating hallucinations is a crucial research of MLLMs.
Hallucination mitigation strategies generally fall into two categories: training-free and training-based methods. Training-free approaches~\citep{huang2024opera,yin2023woodpecker} handle potential hallucinations by post-processing MLLMs' outputs.
On the other hand, training-based approaches aim to reduce hallucinations through instruction fine-tuning~\citep{liu2023mitigating,zhang2024reflective} or preference learning~\citep{gunjal2024detecting,li2023silkie,sun2023aligning,yu2024rlhf,deng2024enhancing,wang2024mdpo}. For example, REVERIE~\citep{zhang2024reflective} is a reflective instruction tuning method that incorporates rationale learning into visual instruction tuning.
As for the preference learning, for example, \citet{gunjal2024detecting} 
proposed Fine-grained DPO (FDPO) and trained a fine-grained multimodal reward model based on InstructBLIP \citep{InstructBLIP}.

Different from previous research, we address the visual-language preference misalignment in MLLMs by introducing a novel cross-modal hierarchical DPO (i.e., CHiP), which simultaneously optimizes preferences in both the text and image modalities from a fine-grained perspective.
Our approach demonstrates better alignment between the two modalities from the visualization of representation and reduced hallucination generation.

\section{Preliminaries}

Direct Preference Optimization (DPO)~\citep{rafailov2024direct} is primarily a preference optimization method that focuses on aligning language models with human preferences without the need for explicit reward modeling or reinforcement learning.
Given a model to be optimized $\pi_{\theta}$, and the reference policy $\pi_{\text{ref}}$, which is a supervised fine-tuning model, the RL optimization of RLHF can be formulated as:
\begin{equation}\label{eq:RL}
\max_{\pi_{\theta}}  \mathbb{E}_{x\sim \mathcal{D}, y\sim \pi_{\theta}(y \mid x)}\bigl[r(x, y)\bigr] - \beta\mathbb{D}_{\textrm{KL}}\bigl[\pi_{\theta}(y\mid x)\mid \mid \pi_{\text{ref}}(y\mid x)\bigr] \,.
\end{equation} 
By maximizing the KL-constrained reward objective to obtain the optimal solution and establishing a mapping between the reward model and the optimal policy, the representation of the reward function is derived as follows:
\begin{equation}
    r(x, y) = \beta \log \frac{\pi_{\theta} (y|x)}{\pi_\text{ref} (y|x)} + \beta \log Z(x) \,,
\end{equation}
where $x$ is the input instruction, $y$ is the response, $\beta$ is a constant, and $Z(x)$ is the partition function.

Given the chosen response $y_w$, where the evaluator preferred it over the rejected response $y_l$,
DPO is expected to learn to maximize the reward difference between chosen ($y_w$) and rejected responses ($y_l$).
The preference optimization objective becomes:
\begin{equation}\label{eq:reward_model}
\begin{aligned}
    \mathcal{L_{DPO}} &= -\mathbb{E}_{(x, y_w, y_l)}\bigl[\log \sigma(r(x, y_w)- r(x, y_l))\bigr] \\
    &=  -\mathbb{E}_{(x, y_w, y_l)}\bigl[\log \sigma(\beta\log \frac{\pi_{\theta} (y_w|x)}{\pi_\text{ref} (y_w|x)}- \beta\log \frac{\pi_{\theta} (y_l|x)}{\pi_\text{ref} (y_l|x)})\bigr] \,,
\end{aligned}
\end{equation}
where DPO learns preferences based on the ranking of the entire response, and the action score can be formulated as:
\begin{equation}
    \log \pi(y|x) = \sum \limits_{y_i\in y} \log p(y_i|x, y_{<i}) \,,
\end{equation}
where $y_i$ denotes the $i$-th token of the response $y$.
During DPO training, the reference model $\pi_\text{ref} (y|x)$ is usually kept fixed while the policy model $\pi_{\theta} (y|x)$ is updated.

\section{Methodology: Cross-modal Hierarchical Direct Preference Optimization}

In this section, we will introduce the Cross-modal Hierarchical  Direct Preference Optimization (CHiP).
CHiP consists of two modules: (1) Hierarchical Textual Preference Optimization, which incorporates preference optimization at the response, segment, and token levels; and (2) Visual Preference Optimization, which addresses the overlooked visual information.

\subsection{Hierarchical Textual Preference Optimization }
Image-based responses are often long and complex, and response-level preference optimization relies on rough rankings of response quality without clearly identifying which segments or tokens contain hallucinations. This makes it challenging to assign credit to desirable behaviors, leading to reward hacking~\citep{laidlaw2024preventing} and the need for a large amount of labeled data. Therefore, we introduce the Hierarchical Textual Preference Optimization module to assign rewards from fine-grained. 

For MLLMs, each sample includes an image ($m$) in addition to prompt $x$, chosen response $y_w$, and rejected response $y_l$. Multimodal DPO relies on both prompt $x$ and image $m$ to select the preferred response from $\{y_w, y_l\}$. 
Next, we will provide a detailed illustration of the three levels of preference optimization for text: response-level, segment-level, and token-level.

\paragraph{Response-level Preference Optimization ($\mathcal{L_{DPO}}_{r})$.}

At the response level, DPO in the MLLMs' scenario aims to maximize $\sigma(r(x, m_w, y_w)- r(x, m_w, y_l))$, and the objective function can be formulated as:
\begin{equation}\label{eq:tpo}
\begin{small}
\begin{aligned}
    \mathcal{L_{DPO}}_{r} =  -\log \sigma \bigg(\beta\log \frac{\pi_{\theta} (y_w|m,x)}{\pi_\text{ref} (y_w|m,x)}- \beta\log \frac{\pi_{\theta} (y_l|m,x)}{\pi_\text{ref} (y_l|m,x)}\bigg) \,,
\end{aligned}
\end{small}
\end{equation}
$\log \pi(y|x)$ can be formulated as:
\begin{equation}\label{eq:rdpo_log}
    \log \pi(y|x, m) = \sum \limits_{y_i\in y} \log p(y_i|x, m, y_{<i}) \,,
\end{equation}
where $y_i$ denotes the $i$-th token of the response $y$.

\paragraph{Segment-level Preference Optimization ($\mathcal{L_{DPO}}_{s}$).}

Intuitively, the corrected segments, particularly entity nouns, play a crucial role in eliminating hallucinations and should be assigned more rewards.
Following ~\cite{yu2024rlhf}, we assign higher rewards to the segments that differ between the chosen response and the rejection response.
Based on \autoref{eq:rdpo_log}, the action score for the segment-level can be denoted as:
\begin{equation}\label{eq:sdpo_log}
\begin{small}
\log \pi^{\text{seg}}(y|x, m) = \frac{1}{C} \bigg( \sum \limits_{y_i\in y} \log p(y_i|x, m, y_{<i}) +  
\gamma \sum \limits_{y_i \in y_c} \log p(y_i|x, m, y_{<i}) \bigg) \,,
\end{small}
\end{equation}
where $y_c$ indicates the segments where changes have occurred. $y_i$ denotes the $i$-th token of the response $y$.
To prevent the model from being misled into giving higher scores to longer responses, $\frac{1}{C}$ is used as a normalization factor, where 
$C = |y| + \gamma * |y_c|$. 
By substituting \autoref{eq:sdpo_log} into \autoref{eq:tpo}, we can obtain the objective function of segment-level preference optimization $\mathcal{L_{DPO}}_{s}$.

\paragraph{Token-level Preference Optimization ($\mathcal{L_{PO}}_{k}$).}
For most previous methods, the optimization objective of DPO was constructed based on sentence-level KL divergence \autoref{eq:RL}). However, the output generated from images is an autoregressive sequence, so aligning MLLMs with human values at the token level is natural. 
Finer-grained alignment not only improves alignment performance but also helps the model maintain diversity~\citep{toekn-dpo-ZengLMYZW24}.
Unlike response-level optimization, which computes a single reward and KL divergence for the entire response, token-level optimization evaluates each token individually, with the cumulative token values forming the score of response.
The sequential KL divergence can be defined as:
\vspace{-6pt}
\begin{equation}
\begin{small}
\begin{aligned}
    \mathcal{L_{PO}}_{k} =& \mathnormal{sg} \left(\beta D_{\mathrm{SeqKL}}\left({x,m},{y}_w;\pi_{\mathrm{ref}}\| \pi_{\theta}\right)\right) - \beta D_{\mathrm{SeqKL}}\left({x,m},{y}_l;\pi_{\mathrm{ref}}\| \pi_{\theta}\right) \,,
\end{aligned}\label{delta2_function}
\end{small}
\end{equation}
where $\mathnormal{sg}$ represents the stop-gradient operator, and
\begin{equation}      
\small{
        D_{\mathrm{SeqKL}}({x,m}, {y};\pi_{\text{ref} }\|\pi_{\theta})=\sum\limits_{t=1}^TD_{\mathrm{KL}}
        (\pi_{\text{ref} }(y|{x}, y^{<t})\|\pi_{\theta}(y|{x}, y^{<t})) \,.
}
\end{equation}
                                                                                                            
\paragraph{Hierarchical Textual Preference Optimization (HDPO).} includes the response, segment, and token-level preference optimization. It can be formulated as:
\begin{equation}\label{eq:hdpo}
\begin{aligned}
    \mathcal{L_{HDPO}} = \mathcal{L_{DPO}}_{r} + \lambda \mathcal{L_{DPO}}_{s}  + \gamma \mathcal{L_{PO}}_{k} \,,
\end{aligned}
\end{equation}
where $\lambda$ and $\gamma$ represent the weights of $\mathcal{L_{DPO}}_{r}$ and $\mathcal{L_{PO}}_{k}$, respectively.

\subsection{Visual Preference Optimization}

To mitigate MLLMs' over-reliance on large language models, we next introduce our Visual Preference Optimization Module. This module forces the model to make preference judgments based on visual information by constructing pairs of images with preferences as variables.
Given a pair of images $(m_w, m_l)$, where $m_w$ allows the prompt $x$ to better match the chosen response $y_w$ compared to $m_l$,
Visual Preference Optimization tries to maximize $\sigma(r(x, m_w, y_w)- r(x, m_l, y_w))$, and the objective function can be formulated as:
\begin{equation}\label{eq:vpo}
\begin{aligned}
\begin{small}
    \mathcal{L_{DPO}}_{v} =  -\log \sigma \bigg(\beta\log \frac{\pi_{\theta} (y_w|m_w, x)}{\pi_\text{ref} (y_w|m_w, x)}- \beta\log \frac{\pi_{\theta} (y_w|m_l, x)}{\pi_\text{ref} (y_w|m_l, x)}\bigg) \,,
\end{small}
\end{aligned}
\end{equation}
where the rejection image $m_l$ can be generated by rotating, cropping, or adding noise to the chosen image $m_w$. 
The objective of CHiP is a combination of the hierarchical textual (\autoref{eq:hdpo}) and visual  (\autoref{eq:vpo}) preference optimizations:
\begin{equation}\label{eq:chip}
\begin{aligned}
    \mathcal{L_{CHiP}} = \mathcal{L_{DPO}}_{v} +
 \mathcal{L_{DPO}}_{r} + \lambda \mathcal{L_{DPO}}_{s}  + \gamma \mathcal{L_{PO}}_{k} \,.
\end{aligned}
\end{equation}

Since the entire response and image encapsulate modality semantics, we assign a weight of 1 (fully consider) to response-level ($\mathcal{L_{DPO}}_{r}$) and visual preference optimization ($\mathcal{L_{DPO}}_{v}$).  $\lambda$ and $\gamma$ ($<1$) (partially consider) adjust the contributions of segment- and token-level preference optimizations.
We refer to the model that only considers response-level and visual preference optimization as Cross-modal Direct Preference Optimization (CMDPO), which can be formulated as:
$
 \mathcal{L_{CMDPO}} = \mathcal{L_{DPO}}_{v} +
 \mathcal{L_{DPO}}_{r}  \,.
 $

Hierarchical textual preference optimization module tries to maximizes $\sigma(r(x, m, y_w)- r(x, m, y_l))$ from different levels,
while the visual preference optimization module tries to maximizes $\sigma(r(x, m_w, y_w)- r(x, m_l, y_w))$. The combination of the them allows MLLMs to choose preferences based on both fine-grained textual and visual modalities.

\section{Experiment and Results}

In this section, we empirically investigate the effectiveness of CHiP in reducing hallucination.

\subsection{Experimental Settings}

\noindent
\textbf{Comparing Models.}
We consider applying CHiP to two different multimodal LLMs: LLaVA-1.6~\citep{liu2024llavanext} and Muffin~\citep{yu2023reformulating}.
For LLaVA-1.6, we have chosen the model size of 7B, using CLIP~\citep{radford@clip} as the visual encoder and Vicuna-1.5-7B~\citep{zheng2023judging} as the LLM backbone.
For Muffin, we have chosen a model size of 13B, using BEiT3~\citep{wang@beit} as the visual module and 13B Vicuna v1.0~\citep{chiang2023vicuna} as the LLM backbone, and a version fine-tuned on the VQAv2 dataset~\citep{goyal2017making} (released by \cite{yu2024rlhf}).

\noindent
\textbf{Training Data.}
There are several publicly available training datasets that include preference pairs for multimodal hallucinations. Here, we choose to use the RLHF-V-Dataset~\citep{yu2024rlhf,yu@rlaif} with 5k training samples as our training dataset.

\noindent
\textbf{Baselines.}
We primarily compare CHiP with standard DPO based on the same models. 
While other multimodal LLMs cannot be directly compared due to differences in base models, preference data, and alignment methods, we provide these results for reference.
LLaVA~\citep{liu2024visual}, 
Muffin~\citep{yu2023reformulating}, 
LRV~\citep{liu2023aligning}, 
LLaVA-RLHF~\citep{sun2023aligning}, 
InstructBLIP~\citep{InstructBLIP}, 
Qwen-VL-Chat~\citep{Qwen-VL},
LLaVA 1.5~\citep{llava15},
RLHF-V~\citep{yu2024rlhf},
HALVA  (13B)~\citep{sarkar2024mitigating}.

\noindent
\textbf{Benchmarks and Evaluation Metrics.}

\begin{itemize*}
    \item \textbf{Object HalBench (ObjHal)}~\citep{rohrbach2018object} is a widely used benchmark for evaluating object hallucination.  
To improve evaluation stability, the benchmark includes 8 diverse prompts and is tested on 300 instances. 
\textbf{Metrics}: Following~\cite{yu2024rlhf,wang2024mdpo}, we report both the \textit{response-level hallucination rate} (\texttt{R.}) and \textit{mention-level hallucination rate} (\texttt{M.}).

    \item \textbf{MMHal-Bench (MMHal)}~\citep{sun2023aligning} is a question-answering benchmark that covers 8 question categories and 12 object topics. 
    \textbf{Metrics:} It uses GPT-4 to assess response quality (\texttt{Ova.}) and hallucination rates (\texttt{R.}).

\item \textbf{HallusionBench}~\citep{guan2024hallusionbench} 
evaluates visual illusions and knowledge hallucinations, featuring 346 images and 1129 questions.
It was the GPT4-assisted evaluation.
\textbf{Metrics:} 
Question Pair Accuracy (\texttt{qA}), 
Figure Accuracy (\texttt{fA}), 
and
All Accuracy (\texttt{aA}). 

\item  
\textbf{AMBER}~\citep{wang2023llm} was designed to be evaluated without LLM assistance. 
Following previous works~\citep{wang2024mdpo}, we only consider the generative tasks.
\textbf{Metrics:} 
(a) CHAIR~\citep{rohrbach2018object} (\texttt{CHAIR});
(b) Object coverage of responses (\texttt{Cover});
(c) Response-level hallucination (\texttt{Hal});
(d) Human cognition hallucination (\texttt{Cog}).
\end{itemize*}

\noindent
\textbf{Implementation Details.}
We train the Muffin (13B)~\citep{yu2023reformulating} and LLaVA-1.6 (7B)~\citep{liu2024llavanext} with CHiP for 3 epochs, with learning rate of 5e-7 and a batch size of 32.
For the training time, LLAVA-1.6 took about three hour to train with CHiP on 4 H100 GPUs, while Muffin took approximately five hours.
Hyperparameter: 
Since our training dataset is RLHF-V dataset~\cite{yu2024rlhf}, we followed~\cite{yu2024rlhf} to set the hyperparameter $\beta=0.5$ and followed \citep{toekn-dpo-ZengLMYZW24} to set $\gamma = 0.1$ for token-level preference optimization. As for the weight of segment-level preference optimization, namely $\lambda$, we set to $\lambda=1$ and $\lambda=3$ for the Muffin and LLava dataset set (\autoref{sec:ablation}).
\textit{How to Identify Hallucinated Segments?} Our training dataset, RLHF-V contains both pre-correction (hallucinated) and post-correction (non-hallucinated) descriptions. We enumerate all segments longer than two tokens in rejected responses and classify those absent in accepted responses as hallucinations. 
\textit{How to construct the rejected images?}
The rejected images are built based on the chosen image of the forward diffusion process at $T=500$ steps (\autoref{sec:noise_steps}).

\begin{table}[htbp]
  \centering \footnotesize
   \caption{The results of hallucination evaluation on the Object HalBench (ObjHal), MMHal-Bench (MMHal), HallusionBench, and AMBER datasets. Values in \textbf{bold} indicate the best performance under the same setting. ``↑'' indicates that a higher value is better for this metric, while ``↓'' indicates that a lower value is better.
      The baseline results are reported in \citep{yu2024rlhf} for ObjHal and MMHal, in \citep{guan2024hallusionbench} for HallucinationBench, and in \citep{wang2024mdpo} for AMBER.
      }
      \vspace{-6pt}
   \renewcommand\tabcolsep{2.4pt}
    \begin{tabular}{lccccccccccc}
    \toprule
    \multirow{2}[4]{*}{Model} & \multicolumn{2}{c}{ObjHal } & \multicolumn{2}{c}{MMHal } & \multicolumn{3}{c}{HallusionBench} & \multicolumn{4}{c}{AMBER } \\
\cmidrule(lr){2-3}\cmidrule(lr){4-5}\cmidrule(lr){6-8}\cmidrule(lr){9-12}          & R.↓ & M.↓ & Ova.↑ & R.↓ & \multicolumn{1}{l}{qA↑} & \multicolumn{1}{l}{fA↑} & \multicolumn{1}{l}{aA↑} & CHAIR↓  & Cover↑ & Hal↓  & Cog↓  \\
    \midrule
    \multicolumn{12}{l}{\textit{Referenced Results (Not Directly Comparable)} }\\
    \midrule
    LLaVA-1.0 \citep{liu2024visual} & 63.0  & 29.5  & -     & 70.8  & -     & -     & -     & -     & -     & -     & - \\
    Muffin \citep{yu2023reformulating} & 50.5  & 24.5  & -     & 68.8  & -     & -     & -     & -     & -     & -     & - \\
    LRV \citep{liu2023aligning}   & 32.3  & 22.3  & -     & 78.1  & 8.8   & 13.0  & 42.8  & -     & -     & -     & - \\
    LLaVA-RLHF \citep{sun2023aligning} & 38.1  & 18.9  & 2.5   & 57.0  & -     & -     & -     & 7.7   & 52.1  & 39.0  & 4.4 \\
    InstructBLIP \citep{InstructBLIP} & 25.9  & 14.3  & 2.1   & 58.0  & 9.5   & 10.1  & 45.3  & 8.8   & 52.2  & 38.2  & 4.4 \\
    Qwen-VL-Chat \citep{Qwen-VL} & 43.8  & 20.0  & 2.9   & 43.0  & 5.9   & 6.7   & 39.2  & 6.6   & 53.2  & 31.0  & 2.9 \\
    LLaVA-1.5 \citep{llava15} & 46.3  & 22.6  & 2.4   & 52.1  & 10.6  & 24.9  & 46.9  & 7.8   & 51.0  & 36.4  & 4.2 \\
    RLHF-V \citep{yu2024rlhf}  & 12.2  & 7.5   & 2.5   & 51.0  & -     & -     & -     & 6.3   & 46.1  & 25.1  & 2.1 \\
    HALVA \citep{sarkar2024mitigating} & -     & -     & -     & -     & 13.9  & 20.1  & 49.1  & -     & -     & -     & - \\
    \rowcolor{gray! 20}GPT-4V \citep{gpt4V}  & 13.6  & 7.3   & -     & 31.3  & 28.8  & 39.9  & 65.3  & 4.6   & 67.1  & 30.7  & 2.6 \\
    \midrule
    Muffin (13B) & 21.5  & 11.6  & 2.4   & 60.42  & 16.0  & 20.8  & 50.9  & 8.0   & \textbf{48.3} & 32.1  & 3.5 \\
    \quad+DPO  & 13.1  & 6.6   & 2.5   & 52.1  & 17.4  & 23.4  & 52.5  & 6.2   & 46.9  & 26.5  & 2.5 \\
    \quad+CHiP & \textbf{6.2} & \textbf{3.9} & \textbf{2.6} & \textbf{49.0} & \textbf{19.1} & \textbf{24.9} & \textbf{54.0} & \textbf{4.4} & 45.3  & \textbf{17.6} & \textbf{1.5} \\
    \midrule
    LLaVA-1.6 (7B) & 14.1  & 7.4   & 2.8   & 42.7  & 15.8  & 20.8  & 51.6  & 8.3   & \textbf{61.0} & 48.6  & 4.2 \\
    \quad+DPO  & 11.0  & 6.6   & 2.7   & 43.8  & 22.2  & \textbf{28.3} & 56.6  & 5.9   & \textbf{61.0} & 38.9  & 3.0 \\
    \quad+CHiP & \textbf{4.9} & \textbf{3.2} & \textbf{2.9} & \textbf{39.6} & \textbf{23.5} & 26.0  & \textbf{58.5} & \textbf{3.7} & 57.8  & \textbf{24.5} & \textbf{1.6} \\
    \bottomrule
    \end{tabular}%
     
  \label{tab:hallu-main}%
\end{table}%

\subsection{Results and Observations}
\autoref{tab:hallu-main} presents the experimental results of applying CHiP to the LLaVA-1.6 and Muffin on four popular hallucination benchmarks.
The main findings are listed below:
(1) \textit{CHiP significantly reduces hallucinations of base models Muffin and LLaVA-1.6.} 
Compared to the base model Muffin (LLaVA), CHiP reduced response- (\texttt{R.}) and mention-levels (\texttt{M.}) hallucinations by 71.2\% (65.3\%) and 66.4\% (56.7\%) relative point on the ObjHal dataset and human cognitive hallucinations (\texttt{Cog}) by 57.1\% (61.9\%) and \text{Hal} by 45.2\% (49.6\%) relative point on the AMBER dataset.
Furthermore, the consistent improvements of CHiP in question pairs (\texttt{qA}) and visual understanding types (\texttt{fA}) on the HallucinationBench, as well as in overall object hallucinations (\texttt{Overall}) on the MMHal dataset.
(2) \textit{Based on Muffin and LLaVA, CHiP consistently outperforms DPO in reducing hallucination on the four benchmarks.}
This indicates that CHiP, which includes the visual preference optimization module and the hierarchical textual preference optimization module, can effectively improve preference alignment performance.
(3) \textit{LLaVA and Muffin with CHiP achieve fewer hallucinations compared to GPT-4 on the ObjHal and AMBER datasets.} 
Compared to GPT-4, LLaVA (Muffin) with CHiP reduced hallucination rates at the response level and mention level by 64.0\% (54.4\%) 
and 56.2\% (46.6\%) relative point respectively on the Object HalBench. On the AMBER dataset, the hallucination rate for the \texttt{Cog} metric decreased by 42.3\% (38.5\%) relative point, with continuous reductions observed across several other categories as well.

\subsection{Ablation Study}
\label{sec:ablation}

\paragraph{Effect of Component Combination.}
To evaluate the contribution of each component in CHiP and the effect of their combinations, we conducted a comprehensive ablation study on CHiP based on LLaVA. 
The experimental results are shown in \autoref{tab:ablation}.
The main observations are as follows: 
(1) \textit{Both hierarchical textual preference optimization (HDPO) and visual preference optimization (CMDPO) are effect.} 
On the ObjHal and MMHal datasets, both HDPO (CHiP-$\mathcal{L_{DPO}}_{v}$) and CMDPO ($\mathcal{L_{DPO}}_{s}$-$\mathcal{L_{PO}}_{t}$) outperform DPO.
This suggests that: (a) more granular preference signals can reduce label ambiguity at the response level (DPO), helping the model learn more effectively; (b) The introduction of visual preference optimization enhances the model's alignment between the image and text.
(2) \textit{The combination of visual preference optimization and hierarchical preference optimization strategies makes DPO the most powerful.}
On both evaluation datasets, CMDPO introduced segment-level ($\mathcal{L_{DPO}}_{s}$), token-level ($\mathcal{L_{PO}}_{t}$), or both segment- and token-levels (CHiP), resulting in a consistent and significant reduction in hallucination rates across different evaluation perspectives, such as response level and mention level. 
Specifically, when CMDPO incorporates both segment-level and token-level optimization (CHiP), the response- and mention-level hallucination rate on ObjHal datasets decreased by 49.6\% and 41.3\%, respectively.

\begin{wrapfigure}{r}{.4\textwidth}
    \vspace{-5mm}
    \centering
    \includegraphics[width=0.81\linewidth]{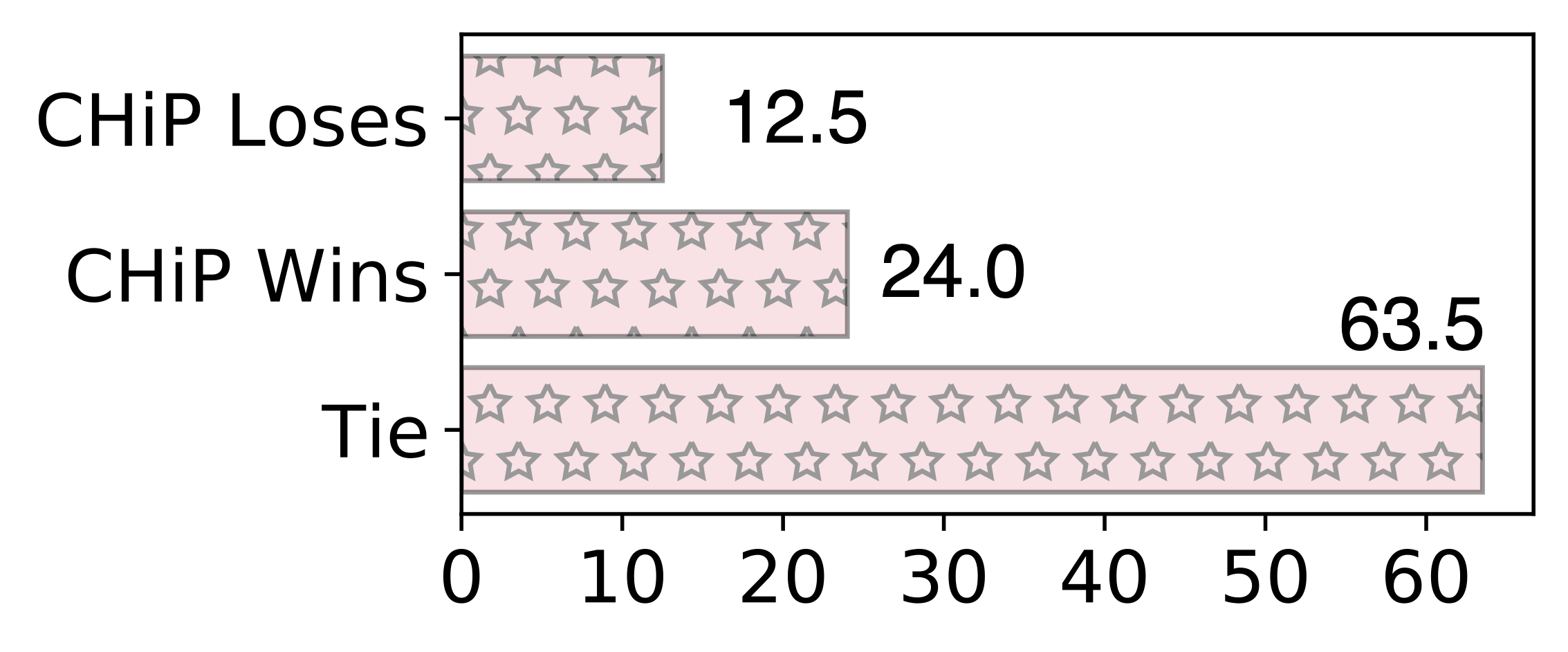}
    \vspace{-8pt}
    \caption{Human evaluation results on MMHal-Bench (MMHal).
    }
 \label{fig:human_mmhal}
\end{wrapfigure}
\paragraph{Human Evaluation.}
Due to incomplete text annotations on the MMHal, GPT-4 couldn't reliably detect hallucinations. To make the results more reliable, we invited experts to manually annotate the data to compare CHiP with DPO based on LLaVA. As shown in Figure \ref{fig:human_mmhal}, CHiP and DPO performed equally on 63.5\% of samples, with CHiP winning 24\%. In the 36.5\% of samples where a distinction was possible, CHiP outperformed DPO in 31.6\%.

\begin{figure}[ht]
	\centering
  \subfigure[Muffin+CHAIR]{
	\begin{minipage}[b]{0.2\textwidth}
	\includegraphics[width=1\textwidth]{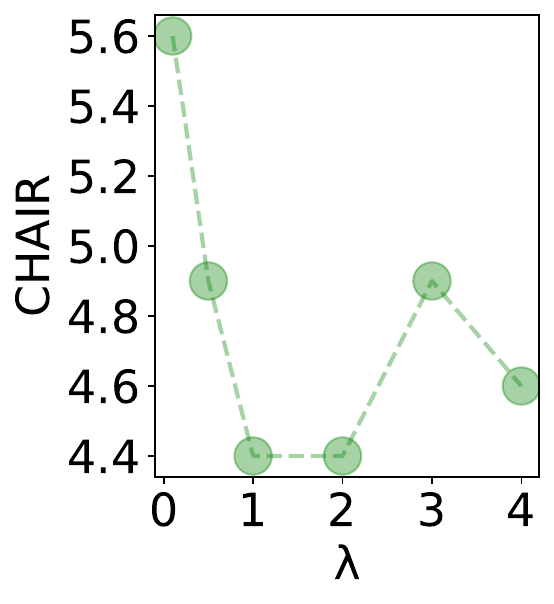}
	\end{minipage}
	}  \hspace{10pt}
    \subfigure[Muffin+Hal Rate]{
	\begin{minipage}[b]{0.2\textwidth}
	\includegraphics[width=1\textwidth]{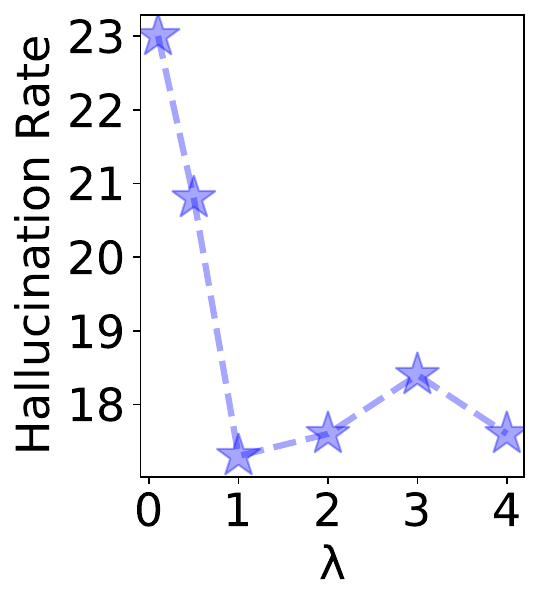}
	\end{minipage}
	}  \hspace{10pt}
 \subfigure[LLaVA+CHAIR]{
	\begin{minipage}[b]{0.2\textwidth}
	\includegraphics[width=1\textwidth]{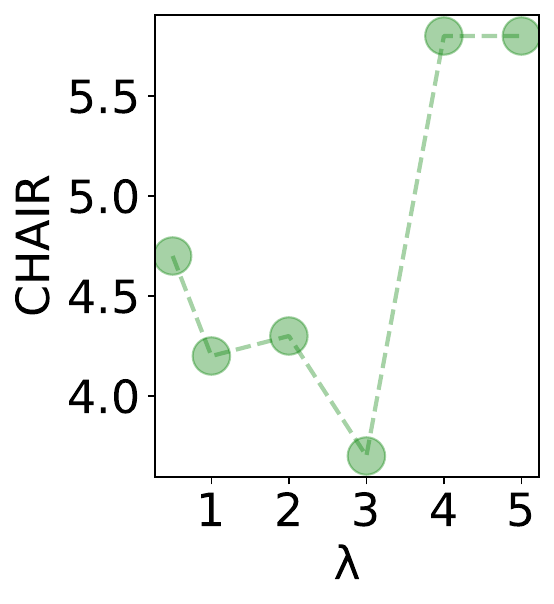}
	\end{minipage}
	}   \hspace{10pt}
    \subfigure[LLaVA+Hal Rate]{
	\begin{minipage}[b]{0.2\textwidth}
	\includegraphics[width=1\textwidth]{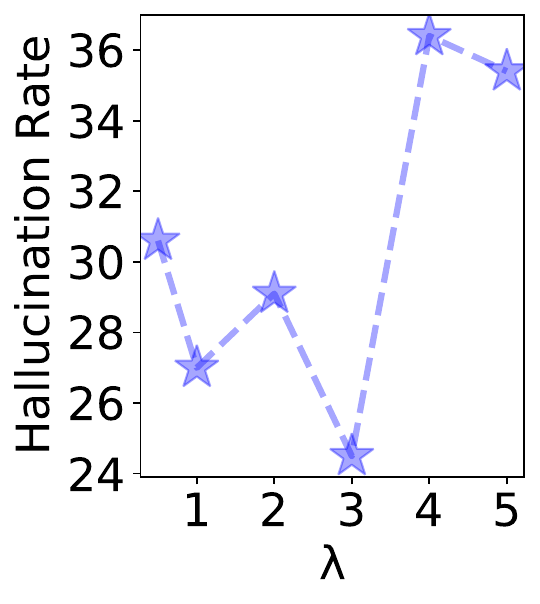}
	\end{minipage}
	}  
    \vspace{-6pt}
	\caption{Results of Muffin+CHiP and LLaVA+CHiP evaluated on the AMBER dataset with different choices of weight $\lambda$ to control the strength of segment-level preference optimization. Findings: When $\lambda = 1$ ($\lambda = 3$), the best performance of the CHAIR and Hallucination Rate metric is achieved on AMBER based on Muffin (LLaVA-1.6).
}
\label{fig:seg_weight}
\end{figure}

\paragraph{Strength of Hierarchical Textual Preference Optimization.}
Hierarchical text preference optimization includes preference optimization at the response, segment, and token levels. Here, we discuss the impact of their weights. 
We fully consider response-level since its global textual semantics by setting its parameter to 1 (\autoref{eq:chip}).
and following~\cite{touvron2023llama}, we set the token-level weight $\gamma = 0.1$. 
As for segment-level, given its crucial role in providing fine-grained feedback on the preference of hallucinations, we fully explore the range of its weight $\lambda$ (as shown in \autoref{eq:hdpo}). 
From the results of \autoref{fig:seg_weight}, we found that the best performance was achieved when $\lambda = 1$ and $\lambda = 3$ for the Muffin and LLaVA frameworks, respectively, and we adopted these settings in all the experiments presented in this paper.

\begin{table}[!htb]
    \begin{minipage}{.4\linewidth}
      \centering  \footnotesize
  \renewcommand\tabcolsep{2pt}
  \caption{The ablation results of CHiP based on LLaVA.  
  Values in \textbf{bold} denote the best performance. 
  }
  \vspace{-6pt}
    \begin{tabular}{lcccc}
    \toprule
    \multirow{2}[2]{*}{Model} & \multicolumn{2}{c}{ObjHal} & \multicolumn{2}{c}{MMHal} \\
    \cmidrule(lr){2-3}\cmidrule(lr){4-5}
          & R.↓ & M.↓ & Ova.↑ & R.↓ \\
    \midrule
    DPO   & 11.03 & 6.61  & 2.73  & 43.75 \\
    \midrule
    CHiP & \textbf{4.92}  & \textbf{3.21}  & \textbf{2.89}  & \textbf{39.63} \\
    -$\mathcal{L_{DPO}}_{v}$  & 9.19  & 5.77  & 2.70  & 42.40 \\
    -$\mathcal{L_{DPO}}_{s}$  & 8.55  & 5.16  & 2.69  & 40.63 \\
    -$\mathcal{L_{PO}}_{t}$  & 6.08  & 3.77  & 2.71  & 40.75 \\
    -$\mathcal{L_{DPO}}_{s}$-$\mathcal{L_{PO}}_{t}$ & 9.76  & 5.47  & 2.78  & 41.71 \\
    \bottomrule
    \end{tabular}%
  \label{tab:ablation}
    \end{minipage}%
    \hspace{8pt}
    \begin{minipage}{.56\linewidth}
      \centering \footnotesize
  \renewcommand\tabcolsep{1pt}
  \caption{Results of training or freezing the visual encoder (VE) in LLaVA during preference optimization.
  $\times$ and \checkmark denote the visual encoder states of training and freezing, respectively. 
  }
  \vspace{-6pt}
   \begin{tabular}{lcccccccc}
    \toprule
    \multicolumn{1}{c}{\multirow{2}[3]{*}{Model}} & \multirow{2}[3]{*}{VE} & \multicolumn{2}{c}{MMHal} & \multicolumn{3}{c}{AMBER} & \multicolumn{2}{c}{ObjHal } \\
\cmidrule{3-9}          &       & Ova.↑ & R.↓ & CHAIR↓  & Cover↑ & Hal↓  & R.↓   & M.↓ \\
    \midrule
    LLaVA & -     & 2.75  & 42.7  & 8.30  & 61.0  & 48.6  & 14.1  & 7.4 \\
    \midrule
    +DPO  & $\times$   & \textbf{2.73} & \textbf{43.8} & 5.94  & 61.0  & 38.9  & 11.0  & 6.6 \\
    +DPO  & \checkmark   & 2.71  & 44.8  & \textbf{5.88} & \textbf{61.6} & \textbf{38.3} & \textbf{10.1} & \textbf{5.7} \\
    \midrule
    +CHiP & $\times$   & \textbf{2.89} & \textbf{39.6} & \textbf{3.72} & \textbf{57.8} & 24.5  & \textbf{4.9} & \textbf{3.2} \\
    +CHiP & \checkmark   & 2.68  & 43.8  & 3.74  & 54.9  & \textbf{22.1} & 5.3   & 3.3 \\
    \bottomrule
    \end{tabular}%
    \label{tab:ve_train}%
    \end{minipage} 
\end{table}

\paragraph{Impact of Training Paradigm.}

The misalignment between image and text semantics is a significant cause of hallucinations in MLLMs~\citep{liu2024survey}. However, most approaches \citep{wang2024mdpo} freeze the visual encoder and train only the connector and LLM during preference optimization.
There raises a scientific question: Can full-model training during MLLM preference optimization reduce hallucinations? 
To investigate this, we explored the impact of freezing versus training the visual backbone on LLaVA enhanced by CHiP and DPO. The results are shown in \autoref{tab:ve_train}. 
\textbf{Results: }
DPO achieves a lower hallucination rate when the visual encoder is trained, whereas CHiP, which incorporates multi-level textual preference and visual preference optimization, does not achieve the expected reduction in hallucination rate when the visual encoder is trained. A possible reason is that the multiple optimization objectives may dilute the model's focus on image-text semantic alignment during the joint training of the visual encoder.

\section{Further Analysis}

\subsection{General Capability Analysis}
Preference learning may compromise a model's general understanding capabilities. In this section, we evaluate and analyze the general capability performance of an MLLM enhanced by our CHiP.
Specifically, we selected several popular general capability evaluation datasets, including MMMU (val)~\citep{yue2024mmmu}, MMMU (test), MMB-ENG~\citep{liu2025mmbench}, MMB-CN, ScienceQA~\citep{lu2022learn}, and LLaVA-Wild~\citep{liu2024visual}. We compared the performance of LLaVA and LLaVA+CHiP on these datasets, with the results shown in \autoref{tab:general}.

\textbf{Observation}: We observe that LLaVA+CHiP outperforms LLaVA on five out of the six datasets. 
This indicates that CHiP slightly improves performance on the MMMU, LLaVA-Wild, and MMB-CN while maintaining comparable performance on others.

\begin{table}[htb]
  \centering \footnotesize
    \caption{The general capability evaluation results. Values in black indicate the best performance, in red show improvement with CHiP, and in green indicate a decline. Values with * are reproduced results. For LLaVA-Wild, we used \textit{gpt-4o-2024-05-13} as evaluator due to \textit{GPT-4-0314} was outdated; for MMMU-test, there was a lack of official LLaVA-1.6 reports.}
  \renewcommand\tabcolsep{4.5pt}
  \vspace{-6pt}
    \begin{tabular}{lcccccc}
    \toprule
          & MMMU(val) & MMMU(test) & MMB-ENG & MMB-CN & ScienceQA & LLaVA-Wild \\
    \midrule
    Num Samples & 900   & 10500 & 6666  & 6666  & 4241  & 90 \\
    LLaVA & 35.80 & 31.70* & \textbf{67.40} & 60.60 & 70.10 & 74.90 \\
    LLaVA+CHiP  & \textbf{36.8$^{\textcolor{magenta}{+1.0}}$} & \textbf{32.1$^{\textcolor{magenta}{+0.4}}$} & 66.6$^{\textcolor{green}{-0.8}}$ & \textbf{60.82$^{\textcolor{magenta}{+0.22}}$} & \textbf{70.15$^{\textcolor{magenta}{+0.05}}$} & \textbf{76.2$^{\textcolor{magenta}{+1.3}}$} \\
    \bottomrule
    \end{tabular}%
  \label{tab:general}%
\end{table}%

\subsection{Impact of Rejection Image Construction Strategy}
\label{sec:noise_steps}
The quality of preference samples is influenced by the quality of the rejection images and the gap between the rejection and chosen images.
In this section, we explore several different strategies for constructing rejection images.

\noindent
\textbf{Strategies.}
The construction rejection images are listed below: 
(1) \textbf{Diffusion}: Following the forward diffusion process in image generation~\citep{diffusion_noise}, small amounts of Gaussian noise are gradually added to the chosen image for T=500 steps. 
(2) \textbf{Blackness}: set all the RGB values of the chosen image to 0.
(3)   \textbf{Crop}: random cropping strategy is utilized to the chosen image.
(4) \textbf{Rotation}: randomly rotate the chosen image by 10 to 80 degrees.
(5) \textbf{Randomness}: select an image from the training set randomly.

\begin{figure}[ht]
	\centering
  \subfigure[Chosen]{
	\begin{minipage}[b]{0.14\textwidth}
	\includegraphics[width=0.9\textwidth]{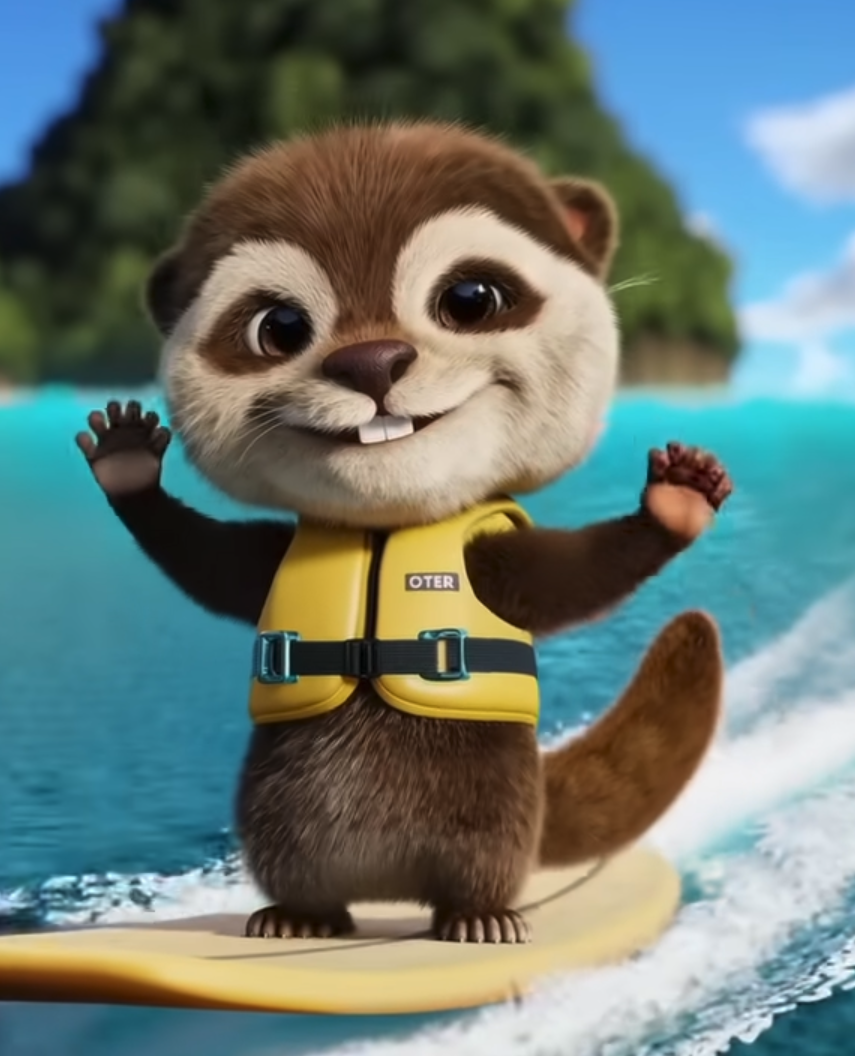}
	\end{minipage}
	}  
    \subfigure[Diffusion]{
	\begin{minipage}[b]{0.14\textwidth}
	\includegraphics[width=0.9\textwidth]{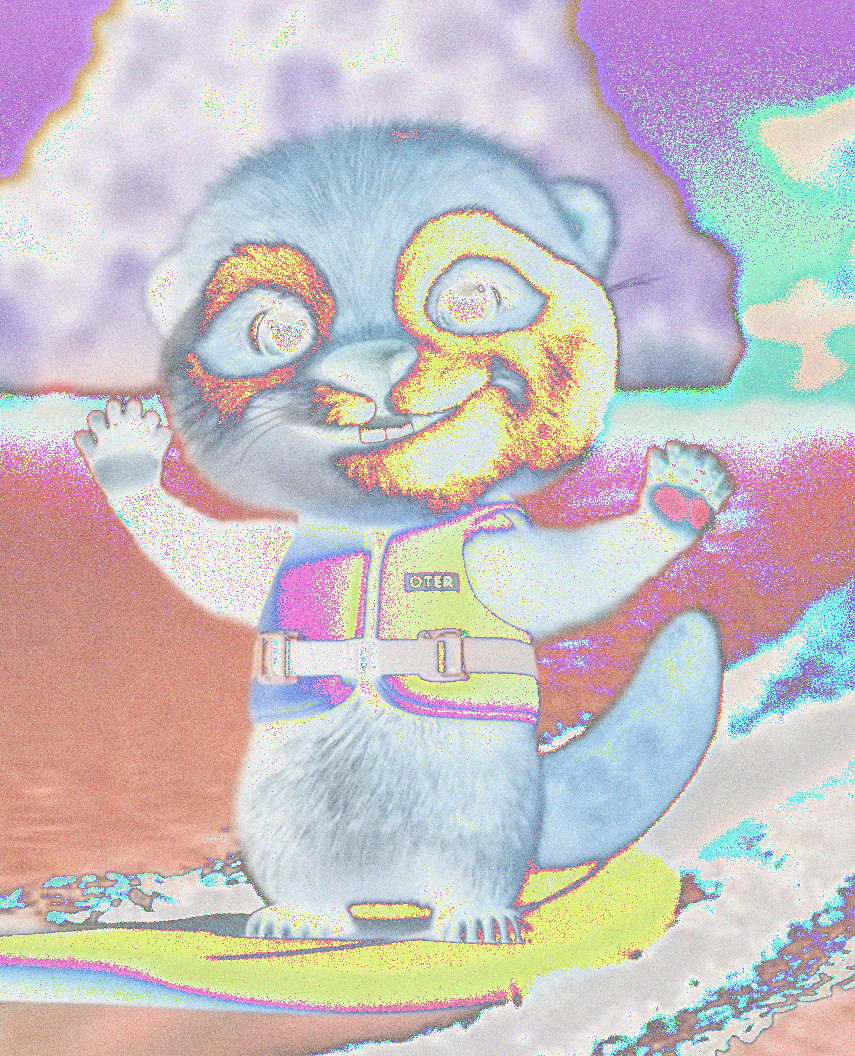}
	\end{minipage}
	}  
	\subfigure[Blackness]{
	\begin{minipage}[b]{0.14 \textwidth}
		\includegraphics[width=0.9\textwidth]{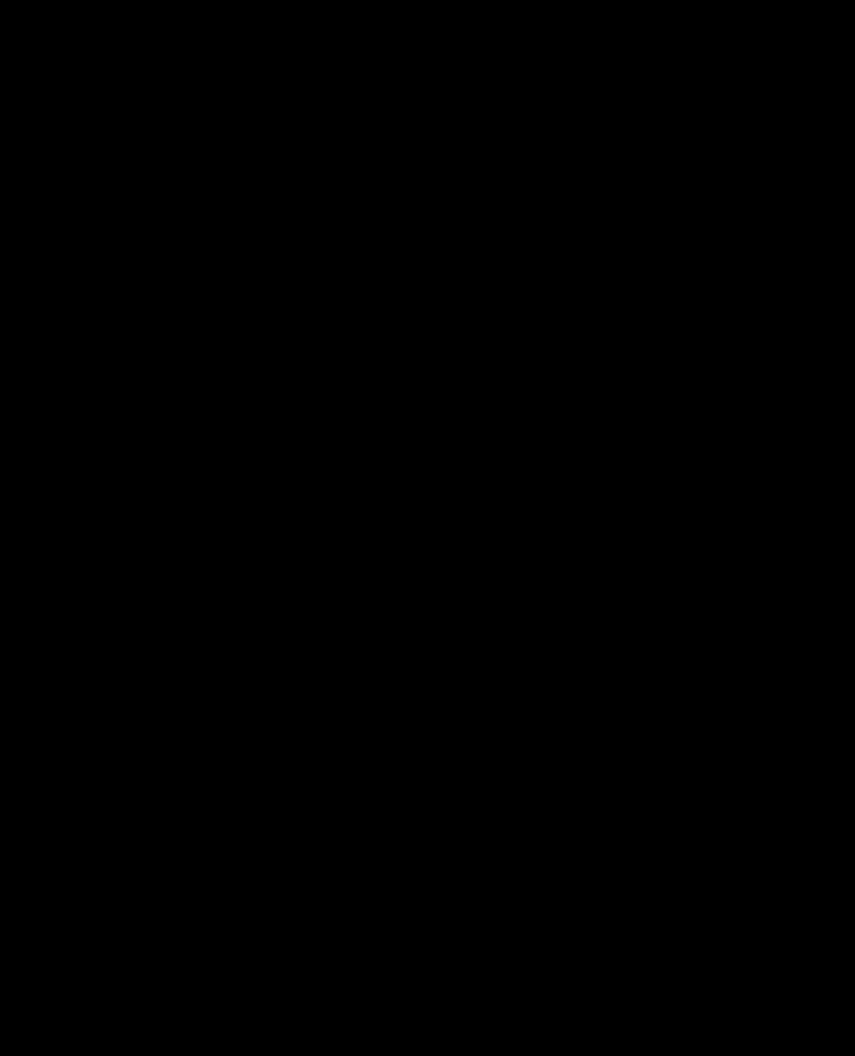}
	\end{minipage}
	} 
	\subfigure[Crop]{
	\begin{minipage}[b]{0.125 \textwidth}
		\includegraphics[width=0.9\textwidth]{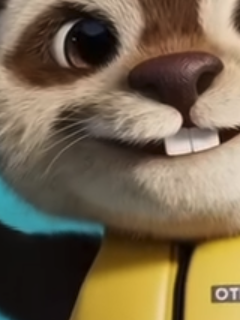}
	\end{minipage}
	} 
         \subfigure[Rotation]{
        	\begin{minipage}[b]{0.14 \textwidth}
        		\includegraphics[width=0.9\textwidth]{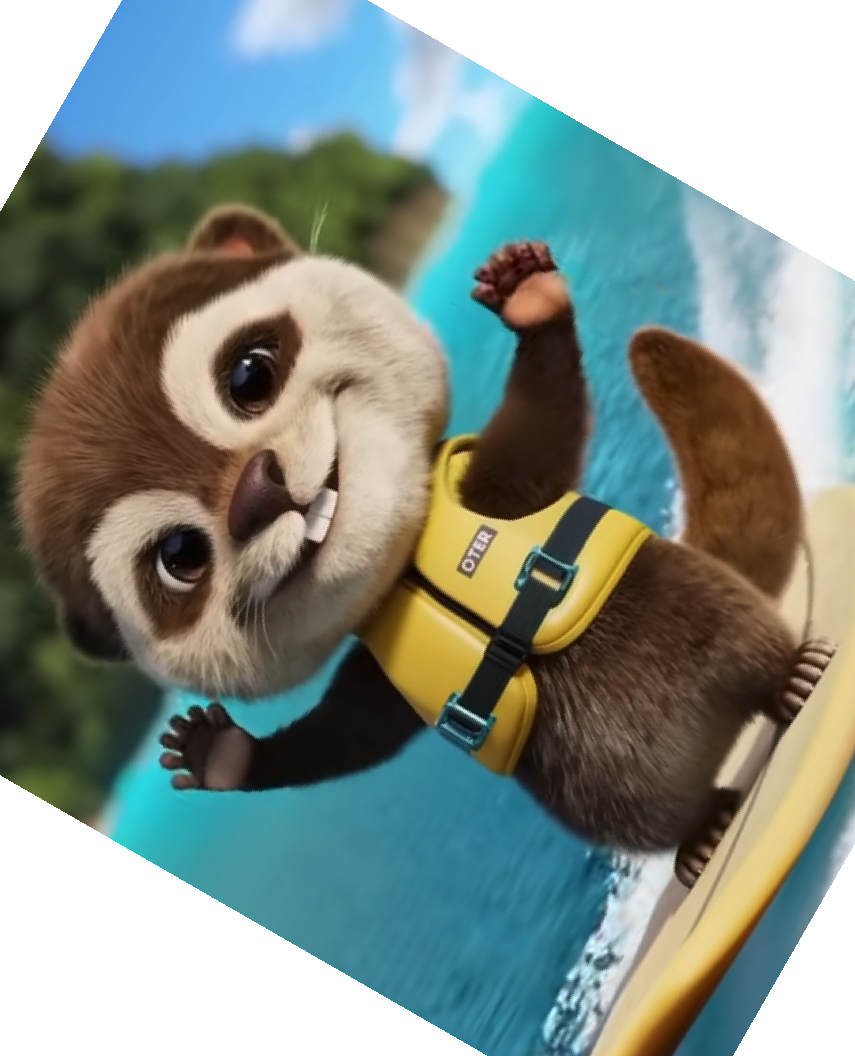}
        	\end{minipage}
        	} 
         \subfigure[Randomness]{
        	\begin{minipage}[b]{0.14 \textwidth}
        		\includegraphics[width=0.9\textwidth]{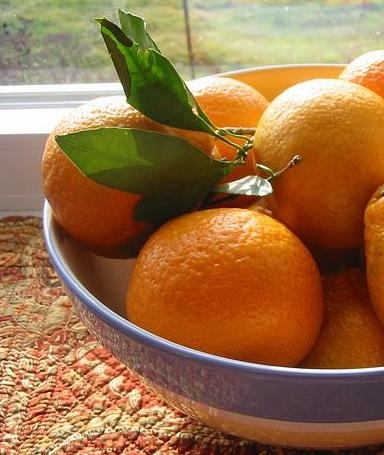}
        	\end{minipage}
        	} 
    \vspace{-8pt}
	\caption{Examples of rejection images constructed by different strategies. (a) is the chosen image.
}
\label{fig:datascale-sft}
\end{figure}

\begin{figure}
\begin{minipage}[b]{.45\linewidth}
    \centering  \footnotesize
\renewcommand\tabcolsep{3.8pt}
\captionof{table}{
Results of CHiP under different rejection image construction strategies. The \textbf{bold} values indicate the best performance. Observation: CHiP achieves the best performance with the diffusion strategy constructed rejection images.
  }
  \vspace{-6pt}
    \begin{tabular}{lcccc}
    \toprule
    \multirow{2}[2]{*}{Strategy} & \multicolumn{2}{c}{ObjHal } & \multicolumn{2}{c}{MMHal } \\
     \cmidrule(lr){2-3}\cmidrule(lr){4-5}
          & R.↓ & M.↓ & Ova.↑ & R.↓ \\
    \midrule
    Diffusion & \textbf{4.9} & \textbf{3.2} & \textbf{2.9} & \textbf{39.6} \\
    Black & 9.4   & 5.0   & 2.4   & 43.8 \\
    Cropping & 5.8   & 3.6   & 2.8   & 40.6 \\
    Random & 10.9  & 5.9   & 2.9   & 41.7 \\
    Rotate & 7.8   & 4.4   & 2.8   & 43.8 \\
    \bottomrule
    \end{tabular}%
    \label{tab:destroy-strategy}%
  \end{minipage}
  \hspace{5pt}
  \begin{minipage}[b]{.53\linewidth}
    \centering 
    \includegraphics[width=0.4\linewidth]{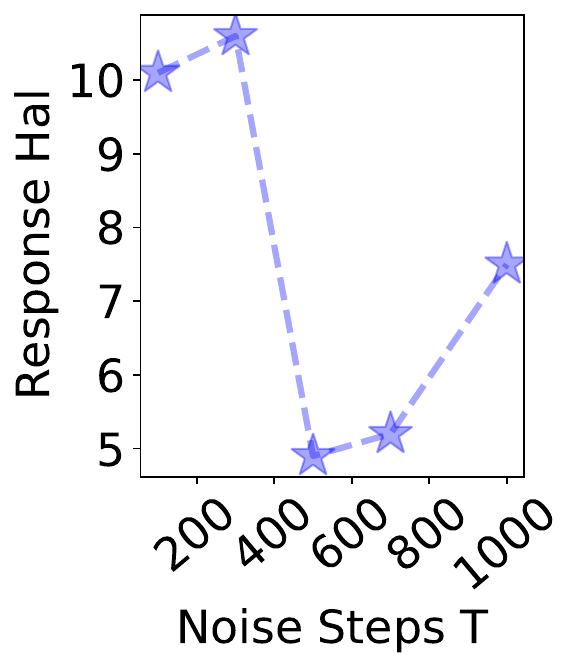} \hspace{10pt}
    \includegraphics[width=0.4\linewidth]{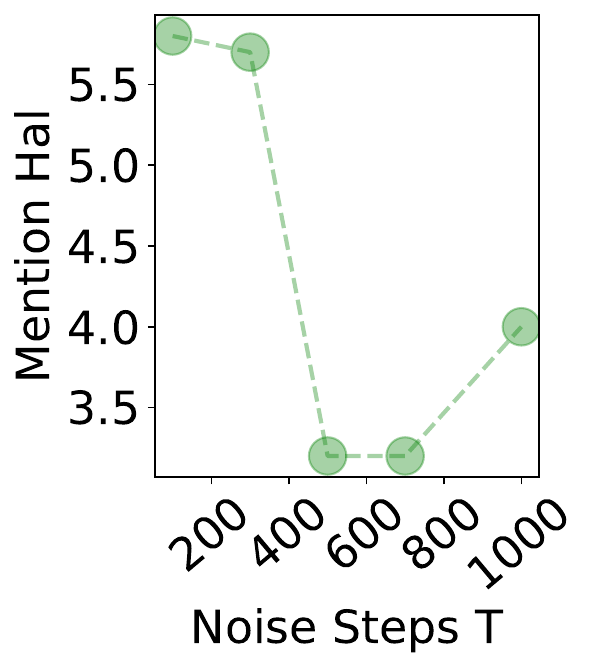}
    \vspace{-7pt}
    \captionof{figure}{Results of LLaVA+CHiP evaluated on the ObjHal dataset with different values of noise step T. ``Response'' represents the response-level hallucination rate, while ``Mention'' represents the mention-level hallucination rate.}%
    \label{fig:diffusion_step}
    \vspace{-40pt}
  \end{minipage}
\end{figure}

\noindent
\textbf{Results.}
The experimental results of CHiP under different construction strategies of rejection images are shown in \autoref{tab:destroy-strategy}.
Observations:
\textbf{The high similarity between the rejection and  chosen images can lead to better preference optimization with CHiP.}
The \texttt{diffusion} and \texttt{cropping} strategies represent blurred and sub-images of the chosen image, respectively, both retaining a significant amount of the chosen image's visual information, resulting in better performance.
However, the \texttt{blackness} and \texttt{randomness} strategies by completely masking and replacing the chosen image almost do not retain the information of the chosen image, leading to poorer performance. Although \texttt{rotation} preserves much of the chosen image's information, it creates significant differences in visual tokens after tokenization, resulting in poor performance.

\paragraph{Impact of Noise Step T.} We further explored the impact of noise steps T on the performance of CHiP, and \autoref{fig:diffusion_step} shows the performance curve on the ObjHal dataset. 
\underline{CHiP achieves the best performance when the noise steps T=500.} The possible reasons are: (1) Fewer noise steps make the rejection image too similar to the chosen image, causing preference label ambiguity and reducing the effectiveness of the visual preference optimization module. (2) Too many noise steps erase most of the image's information, making it easily distinguishable, which hinders the module's ability to leverage the visual modality.

\subsection{Representation Visualization}
\begin{figure}[t]
	\centering
    \subfigure[LLaVA]{
	\begin{minipage}[b]{0.2\textwidth}
	\includegraphics[width=1\textwidth]{fig/pca/pca_llava.pdf}
	\end{minipage}
	}  \hspace{5pt}
	\subfigure[LLaVA+DPO]{
	\begin{minipage}[b]{0.2 \textwidth}
		\includegraphics[width=1\textwidth]{fig/pca/pca_dpo.pdf}
	\end{minipage}
	} \hspace{5pt}
    \subfigure[LLaVA+CMDPO]{
	\begin{minipage}[b]{0.2 \textwidth}
		\includegraphics[width=1\textwidth]{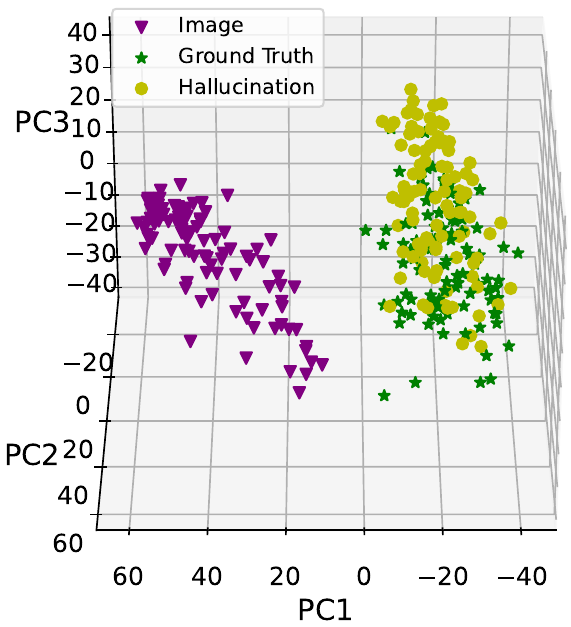}
	\end{minipage}
	} \hspace{5pt}
	\subfigure[LLaVA+CHiP]{
	\begin{minipage}[b]{0.2 \textwidth}
		\includegraphics[width=1\textwidth]{fig/pca/pca_hcmdpo.pdf}
	\end{minipage}
	} 
        \vspace{-6pt}
	\caption{The visualization of representation distributions of various preference optimization strategies. Findings: our CHiP makes more alignment between images and non-hallucination descriptions and improves the model's ability to distinguish between hallucinatory and non-hallucinatory text.
}
\label{fig:visualization}
\end{figure}

Ideally, a MLLM with a low hallucination rate should ensure that the representations of the textual description for the image are as close as possible to the representations of the image itself, while keeping the representations of hallucinated and non-hallucinated texts as far apart as possible.
We analyze the effectiveness of CHiP compared to DPO and several ablation preference optimization strategies from a representational perspective. Specifically, we sampled 150 image-ground truth description pairs from the COCO-2017~\citep{lin2014microsoft} validation set, and GPT-4 was used to generate more detailed non-hallucinated descriptions (manually verified), as well as hallucinated descriptions. We then input the images, non-hallucinated texts, and hallucinated texts into LLaVA separately, and took the representation of the last token of the LLM (in this case, LLaMA) as the representation of the text or the image. We applied Principal Component Analysis (PCA)~\footnote{\url{https://en.wikipedia.org/wiki/Principal_component_analysis}} to reduce the dimensionality of the high-dimensional representations, and the results are shown in \autoref{fig:visualization}.

\textbf{Observations:} Compared to original LLaVA, while DPO still exhibits a disconnect between the representations of image and text modalities, it is able to differentiate between hallucinated and non-hallucinated texts. After introducing image preference optimization based on DPO, namely CMDPO, the model not only distinguishes between hallucinated and non-hallucinated texts but also brings the representations of the image and the ground-truth description closer. With the introduction of more fine-grained text and image preference optimization, namely CHiP, the alignment between the image and ground-truth descriptions becomes even closer, while maintaining the ability to distinguish between hallucinated and non-hallucinated texts.

\section{Conclusion}
In this paper, we tackle the issue of hallucinations in multimodal large language models (MLLMs) by proposing Cross-Modal Hierarchical DPO (CHiP). CHiP integrates visual and hierarchical textual preference optimization, facilitating cross-modal preference capture and finer-grained distinctions. Experiments on four widely-used datasets demonstrate that CHiP effectively reduces hallucinations.
We visualized the representations of images, non-hallucinatory descriptions, and hallucinatory descriptions. The results show that CHiP, compared to standard multimodal DPO, more effectively bridges the semantic gap between images and non-hallucinatory descriptions while enhancing the distinction between hallucinatory and non-hallucinatory descriptions.


\newpage
\bibliography{iclr2025_conference}
\bibliographystyle{iclr2025_conference}

\newpage
\appendix

\section{Differences between Token- and Response-level Optimization}
Here, we first summarize the differences between token-level optimization and response-level optimization (in \autoref{sec:overview}). For clarity, the following two subsections provide detailed derivations of response-level (in \autoref{sec:response})  and token-level (in \autoref{sec:token}) optimization for reference.

\subsection{Overview}
\label{sec:overview}
The preference optimization function can be divided into two components: (1) the reward function, which quantifies user preferences and drives the optimization direction; and (2) KL divergence, which controls the difference between the output distributions of the policy model and the preference model.

Given the input instruction $x$, image $m$, and response $y$, the \textbf{Response-level Preference Optimization} can be formulated as below:
\begin{equation}\label{eq:RL2-samll}
\max_{\pi_{\theta}}  \mathbb{E}_{x,m\sim \mathcal{D}, y\sim \pi_{\theta}(y \mid x)}\bigl[r(x,m, y)\bigr] - \beta\mathbb{D}_{\textrm{KL}}\bigl[\pi_{\theta}(y\mid x,m)\mid \mid \pi_{\text{ref}}(y\mid x,m)\bigr] ,
\end{equation} 
where the reward function can be denoted as:
\begin{equation}\label{eq:reward-resp}
    r(x,m, y) = \beta \log \frac{\pi_{\theta} (y|x,m)}{\pi_\text{ref} (y|x,m)} + \beta \log Z(x,m) \,,
\end{equation}

And, the \textbf{Token-level Preference Optimization} can be defined as below:
\begin{equation}\label{eq:token}
\begin{split}\small
\max_{\pi_{\theta}} \ \mathbb{E}_{{x,m}, {y}^{<t}\sim\mathcal{D},y^t\sim \pi_{\theta}(\cdot|[{x,m},y^{<t}])} 
\bigl[ r(x,m, y) \bigr] 
-D_{\mathrm{SeqKL}}({x,m}, {y};\pi_{\theta}\|\pi_{ref})),
    \end{split}
\end{equation}
where the reward function can be denoted as:
\begin{equation}\label{eq:reward-token}
    r(x,m, y) =  \sum_{t=1}^T\gamma^{t-1}R([{x,m},y^{<t}], y^t)
\end{equation}\label{reward_tok}
where $ D_{\mathrm{SeqKL}}({x,m}, {y};\pi_{\theta}\|\pi_{ref})$ denotes the sequential KL divergence, and it can be defined as:
\begin{equation}\label{eq:kl_token}
\begin{split}\small
    D_{\mathrm{SeqKL}}({x,m}, {y};\pi_{\theta}\|\pi_{ref})=
    \sum\limits_{t=1}^TD_{\mathrm{KL}}(\pi_{\theta}(\cdot|[{x,m}, y^{<t}])\|\pi_{ref}(\cdot|[{x,m}, y^{<t}])).
\end{split}
\end{equation}

\paragraph{Comparison: }
\paragraph{(1) Reward Function}
For response-level preference optimization, the reward function is calculated based on the generation probabilities of a response by the policy model and the reference model (as shown in \autoref{eq:reward-resp}). The probability of the entire response is first obtained, and then the reward for the entire response is calculated.

For token-level preference optimization, the reward function calculates the reward for each token individually (for example, $y^{t}$, based on the $x$ and $m$, and $y^{<t}$) and then sums up the rewards of all tokens to obtain the reward for the entire response, as shown in \autoref{eq:reward-token}.

\paragraph{(2) KL Divergence}
For response-level optimization (as shown in \autoref{eq:RL2-samll}), KL divergence is calculated as the distance between the distributions of the response $y$ given $x$ and $m$, as modeled by the policy model and the reference model (note: this is based on the response distribution).

\noindent
For token-level optimization, KL divergence is computed as the distance between the distributions of $y^{t}$ given $x$ and $m$, and $y^{<t}$, as modeled by the policy model and the reference model. The overall distance between the policy and reference models is obtained by summing these distances across all tokens. \autoref{eq:kl_token} formulate this process.

\subsection{Response-level Preference Optimization}
\label{sec:response}
Direct Preference Optimization (DPO)~\citep{rafailov2024direct} is primarily a preference optimization method that focuses on aligning language models with human preferences without the need for explicit reward modeling or reinforcement learning.
Given a model to be optimized $\pi_{\theta}$, and the reference policy $\pi_{\text{ref}}$, which is a supervised fine-tuning model, the RL optimization of RLHF can be formulated as:
\begin{equation}\label{eq:RL2}
\max_{\pi_{\theta}}  \mathbb{E}_{x,m\sim \mathcal{D}, y\sim \pi_{\theta}(y \mid x)}\bigl[r(x,m, y)\bigr] - \beta\mathbb{D}_{\textrm{KL}}\bigl[\pi_{\theta}(y\mid x,m)\mid \mid \pi_{\text{ref}}(y\mid x,m)\bigr] \,.
\end{equation} 
By maximizing the KL-constrained reward objective to obtain the optimal solution and establishing a mapping between the reward model and the optimal policy, the representation of the reward function is derived as follows:
\begin{equation}\label{eq:reward}
    r(x,m, y) = \beta \log \frac{\pi_{\theta} (y|x,m)}{\pi_\text{ref} (y|x,m)} + \beta \log Z(x,m) \,,
\end{equation}
where $x$ is the input instruction, $m$ is the image, $y$ is the response, $\beta$ is a constant, and $Z(x,m)$ is the partition function.

Given the chosen response $y_w$, where the evaluator preferred it over the rejected response $y_l$,
DPO is expected to learn to maximize the reward difference between chosen ($y_w$) and rejected responses ($y_l$).
The preference optimization objective becomes:
\begin{equation}
\begin{aligned}\label{eq:obj}
    \mathcal{L_{DPO}} &= -\mathbb{E}_{(x,m, y_w, y_l)}\bigl[\log \sigma(r(x,m, y_w)- r(x,m, y_l))\bigr] \\
    &=  -\mathbb{E}_{(x,m, y_w, y_l)}\bigl[\log \sigma(\beta\log \frac{\pi_{\theta} (y_w|x,m)}{\pi_\text{ref} (y_w|x,m)}- \beta\log \frac{\pi_{\theta} (y_l|x,m)}{\pi_\text{ref} (y_l|x,m)})\bigr] \,,
\end{aligned}
\end{equation}
where DPO learns preferences based on the ranking of the entire response, and the action score can be formulated as:
\begin{equation}
    \log \pi(y|x,m) = \sum \limits_{y_i\in y} \log p(y_i|x,m, y_{<i}) \,,
\end{equation}
where $y_i$ denotes the $i$-th token of the response $y$.
During DPO training, the reference model $\pi_\text{ref} (y|x,m)$ is usually kept fixed while the policy model $\pi_{\theta} (y|x,m)$ is updated.

\subsection{Token-level Preference Optimization}
\label{sec:token}

The objective function of DPO operates at the sentence level, as shown in \autoref{eq:RL2}. The principle of token-level preference optimization is similar to sentence-level preference optimization. The difference between them lies in the reward function. In token-level preference optimization, the reward function is token-wise, which can be viewed as the cumulative reward for generating the text.

Given a response composed of $T$ tokens ${y} =  [y^1, y^2, ..., y^T]$, where $y^t\in \mathcal{Y}$, and $\mathcal{Y}$ represents the vocabulary. 
Additionally, we define $y^{<1}=[\ ]$.
Given a prompt ${x}$, a image $m$, and model-generated response ${y}$'s first $t-1$ tokens,  the LM predicts the probability distribution of the next token can be formulated as $\pi_{\theta}(\cdot|[{x,m}, y^{<t}])$. 
Therefore, the objective of token-level preference optimization can be denoted as below:

\begin{equation}\label{eq:token}
\begin{split}
\max_{\pi_{\theta}} \ \mathbb{E}_{{x,m}, {y}^{<t}\sim\mathcal{D},y^t\sim \pi_{\theta}(\cdot|[{x,m},y^{<t}])} 
\bigl[ \sum_{t=1}^T\gamma^{t-1}R([{x,m},y^{<t}], y^t) \bigr] 
-D_{\mathrm{SeqKL}}({x,m}, {y};\pi_{\theta}\|\pi_{ref})),
    \end{split}
\end{equation}
where $ D_{\mathrm{SeqKL}}({x,m}, {y};\pi_{\theta}\|\pi_{ref})$ denotes the sequential KL divergence, and it can be defined as:
\begin{equation}
\begin{split}
    D_{\mathrm{SeqKL}}({x,m}, {y};\pi_{\theta}\|\pi_{ref})=
    \sum\limits_{t=1}^TD_{\mathrm{KL}}(\pi_{\theta}(\cdot|[{x,m}, y^{<t}])\|\pi_{ref}(\cdot|[{x,m}, y^{<t}])).
\end{split}
\end{equation}
where $\sum_{t=1}^T\gamma^{t-1}R([{x,m},y^{<t}], y^t)$ is the accumulate reward, and $\gamma$ represents a weight and is a constant.

We can follow a similar approach to response-level preference optimization by maximizing the KL-constrained reward objective to obtain the optimal solution, thereby deriving a reward function similar to \autoref{eq:reward}.
And further derive the token-level preference optimization function like sentence-level shown in \autoref{eq:obj}.
However, $Z([{x,m},y_w^{<t}];\beta)\ne Z([{x,m},y_l^{<t}];\beta)$, which means that optimization at the sentence-level preference pairs can result in the cancellation of policies, while the cancellation does not occur in token-level preference optimization.

Here, we directly employ the Bradley-Terry model to represent the probability of human preferences based on the optimal policy. In the KL-constrained advantage maximization problem associated with \autoref{eq:token}, the Bradley-Terry model takes the optimal policy $\pi_{\theta}$ and the reference policy $\pi_{\mathrm{ref}}$ to expresses human preference probabilities:

\begin{equation}
    \small
        P_{\mathrm{BT}}({y}_w \succ {y}_l |{x,m})=\sigma(\lambda({x,m}, {y}_w, {y}_l) - \delta({x,m}, {y}_w, {y}_l)),\label{PBT_pi}
    \end{equation}
    where, $\lambda({x,m}, {y}_w, {y}_l)$ refers to the difference in rewards implicitly defined by the language model $\pi_{\theta}$ and the reference model $\pi_{\mathrm{ref}}$ \citep{rafailov2024direct}, represented as
    \begin{equation}
    \small
    \lambda({x,m}, {y}_w, {y}_l)=\beta\log\frac{\pi_{\theta}({y}_w\mid {x,m})}{\pi_{\mathrm{ref}}({y}_w\mid {x,m})}-\beta\log\frac{\pi_{\theta}({y}_l\mid {x,m})}{\pi_{\mathrm{ref}}({y}_l\mid {x,m})}, \label{u_function}
    \end{equation}
    and $\delta({x,m}, {y}_w, {y}_l)$ is the difference in sequential forward KL divergence between the preference pairs $({x,m}, {y}_w)$ and $({x,m}, {y}_l)$.
    \begin{equation}
        \begin{aligned}
    \delta({x,m}, {y}_w, {y}_l) =\beta D_{\mathrm{SeqKL}}\left({x,m},{y}_w;\pi_{\mathrm{ref}}\| \pi_{\theta}\right) -\beta D_{\mathrm{SeqKL}}\left({x,m},{y}_l;\pi_{\mathrm{ref}}\| \pi_{\theta}\right),
        \end{aligned}\label{delta_function}
    \end{equation}
where $\beta$ is the weight. Put them together, we obtain the loss function for token-level preference optimization:

\begin{equation}
\begin{aligned}
    \mathcal{L}_{\mathrm{TDPO}} &= -\mathbb{E}_{(x, m, y_w, y_l)} \bigg[\log\sigma\bigg(\lambda(x, m, y_w, y_l) - \delta(x, m, y_w, y_l)\bigg)\bigg], \\
    &= -\mathbb{E}_{(x, m, y_w, y_l)} \bigg[\log\sigma\bigg( 
        \beta \log\frac{\pi_{\theta}(y_w \mid x, m)}{\pi_{\mathrm{ref}}(y_w \mid x, m)} 
        - \beta \log\frac{\pi_{\theta}(y_l \mid x, m)}{\pi_{\mathrm{ref}}(y_l \mid x, m)} \\
    &\quad - \alpha \Bigl(
        \beta D_{\mathrm{SeqKL}}\bigl(x, m, y_2; \pi_{\mathrm{ref}} \| \pi_{\theta}\bigr) 
        - \mathnormal{sg}\bigl(\beta D_{\mathrm{SeqKL}}(x, m, y_1; \pi_{\mathrm{ref}} \| \pi_{\theta})\bigr)
    \Bigr)
    \bigg)\bigg],
\end{aligned}
\end{equation}
\paragraph{Token-level Preference Optimization ($\mathcal{L_{PO}}_{k}$).}
The difference in reward $\lambda({x,m}, {y}_w, {y}_l)$ appears in both the response-level and segment-level preference optimizations in this work. Therefore, we consider solely on sequential KL divergence, which serves as the optimization term for token-level preference optimization:
\begin{equation}
\begin{aligned}
    \mathcal{L_{PO}}_{k} =& \mathnormal{sg} \left(\beta D_{\mathrm{SeqKL}}\left({x,m},{y}_w;\pi_{\mathrm{ref}}\| \pi_{\theta}\right)\right) - \beta D_{\mathrm{SeqKL}}\left({x,m},{y}_l;\pi_{\mathrm{ref}}\| \pi_{\theta}\right) \,,
\end{aligned}\label{delta2_function}
\end{equation}
where $\mathnormal{sg}$ represents the stop-gradient operator, and
\begin{equation}                                      
        D_{\mathrm{SeqKL}}({x,m}, {y};\pi_{\text{ref} }\|\pi_{\theta})=\sum\limits_{t=1}^TD_{\mathrm{KL}}
        (\pi_{\text{ref} }(y|{x,m}, y^{<t})\|\pi_{\theta}(y|{x,m}, y^{<t})) \,.
\end{equation}

\section{Further Analysis}

\subsection{Does CHiP make the model less talkative?}
\textbf{Why does CHiP lower performance on the metric of \texttt{Cover} (Object coverage of responses)?}
The results in \autoref{tab:hallu-main} show that while CHiP reduces the hallucination rate, it also decreases object coverage on the AMBER dataset. This raises a question: does CHiP reduce hallucinations by limiting the amount of text generated?
To answer this question, we calculated the average output length of LLaVA, LLaVA+DPO, and LLaVA+CHiP across three generative datasets, namely AMBER, MMHal, and ObjHal. Since HallusionBench is a multiple-choice task, calculating response length is not meaningful, so we have omitted it here.
The results are presented in \autoref{tab:avg-length}. We can observe that the output lengths of the three models are comparable. Therefore, CHiP lowers the hallucination rate without making the model less talkative. Furthermore, a manual analysis of LLaVA+CHiP's responses reveals that when an image contains ambiguous or uncertain objects or attributes, CHiP tends to omit to mention them, effectively reducing hallucinations.

\textbf{Why does CHiP lower performance in \texttt{fA} (Figure Accuracy)?}

The \texttt{fA} metric is designed to evaluate the model's logical consistency, specifically ensuring that responses to questions are not based on random guesses. From \autoref{tab:hallu-main}, we observe that on LLaVA-7B, CHiP achieves a lower \texttt{fA} compared to DPO, but higher than the original LLaVA-7B. A possible reason is that DPO directly targets response-level preference optimization, focusing on aligning the model's outputs with human preferences at the response level, which makes it better at maintaining logical consistency. In contrast, CHiP's optimization objectives include segment-level, token-level, and image preference optimization. This additional complexity may dilute the focus on logical consistency during the optimization process, resulting in CHiP's slightly lower \texttt{fA} compared to DPO.

\begin{table}[htb]
  \centering
  \caption{Average output length statistics of different models.}
    \begin{tabular}{lccc}
    \toprule
    Model & AMBER & MMHal & ObjHal \\
    \midrule
    LLaVA & 131   & 44    & 164 \\
    LLaVA+DPO & 127   & 40    & 160 \\
    LLaVA+CHiP & 124   & 43    & 151 \\
    \bottomrule
    \end{tabular}%
  \label{tab:avg-length}%
\end{table}%

\subsection{Annotator Background in Human Evaluation}
 To mitigate the errors in GPT evaluations caused by incomplete annotations in the MMHal dataset, we introduced human experts for manual evaluation to ensure more reliable conclusions: CHiP has a lower hallucination rate compared to traditional DPO. Details about the annotators: We invited three human experts who specialize in multimodal hallucination tasks. These experts are well-versed in various types of hallucinations, such as object hallucinations, attribute hallucinations, and environment hallucinations. We consider object hallucinations to be more severe than attribute and environment hallucinations. This prioritization allows the expert to efficiently classify and label hallucination types during the annotation process, enabling them to assign reasonable and accurate scores.

\subsection{Effect of Token-level Preference Optimization}
In this section, we explored whether using token-level preference optimization independently as an optimization objective would still allow the model to work effectively. The evaluation results on four datasets are presented in the \autoref{tab:effect-tdpo} It can be observed that solely using token-level preference optimization results in lower hallucination rates compared to traditional DPO and the original LLaVA, which holds for the four datasets.

\begin{table}[htb]
  \centering \footnotesize
  \renewcommand\tabcolsep{2.4pt}
  \caption{Performance comparison of LLaVA after token-level preference optimization (TDPO) and direct preference optimization (DPO). Values in bold indicate the best performance.}
    \begin{tabular}{lccccccccccc}
    \toprule
    \multirow{2}[2]{*}{Model} & \multicolumn{2}{c}{ObjHal} & \multicolumn{2}{c}{MMHal} & \multicolumn{3}{c}{HallusionBench} & \multicolumn{4}{c}{AMBER} \\
    \cmidrule(lr){2-3}\cmidrule(lr){4-5}\cmidrule(lr){6-8}\cmidrule(lr){9-12}
          & Resp.↓ & Mention↓ & Overall↑ & Resp.↓ & (qAcc) ↑ & (fAcc)↑ & (aAcc)↑ & CHAIR↓ & Cover↑ & Hal↓  & Cog↓ \\
    \midrule
    TDPO  & \textbf{9.56} & \textbf{5.50} & \textbf{2.73} & \textbf{42.71} & \textbf{22.64} & \textbf{28.32} & \textbf{57.22} & \textbf{5.90} & \textbf{61.30} & \textbf{37.60} & 3.10 \\
    DPO   & 11.03 & 6.61  & 2.73  & 43.75 & 22.20 & 28.32 & 56.60 & 5.90  & 61.00 & 38.90 & \textbf{3.00} \\
    \bottomrule
    \end{tabular}%
  \label{tab:effect-tdpo}%
\end{table}%

\subsection{Fine-grained Analysis}

Here, we conducted a fine-grained evaluation of the base model (i.e., Muffin and LLaVA), DPO, and CHiP on the \textbf{AMBER}, \textbf{MMHal-Bench (MMHal)}, and \textbf{Object HalBench (ObjHal)} datasets. The results are shown in \autoref{tab:fine_amber}, \autoref{tab:fine_mmhal}, and \autoref{tab:fine_objhal}. The main finding is that CHiP performed well on many fine-grained evaluation metrics across these three datasets.

\begin{table}[htbp]
  \centering \footnotesize
  \renewcommand\tabcolsep{2.4pt}
  \caption{Fine-grained results on \textbf{AMBER}. Bold values indicates the best performance. ↑ indicates that a higher value represents better performance, while ↓ indicates that a lower value is better. Findings: Our CHiP achieves the best \texttt{AMBER Score} under both the base model Muffin and the LLaVA model.}
    \begin{tabular}{lccccccccc}
    \toprule
    \multirow{2}[2]{*}{Model} & \multicolumn{4}{c}{Generative } & \multicolumn{4}{c}{Discriminative} & AMBER \\
    \cmidrule(lr){2-5}\cmidrule(lr){6-9}
          & CHAIR↓  & Cover↑ & Hal↓  & Cog↓  & F1↑   & F1E↑  & F1A↑  & F1R↑  & Score↑ \\
    \midrule
    Muffin (13B) & 8.0   & \textbf{48.3} & 32.1  & 3.5   & 86.4  & 95.0  & 79.3  & 71.4  & 89.20 \\
    +DPO  & 6.2   & 46.9  & 26.5  & 2.5   & 86.9  & 95.9  & 79.9  & 70.4  & 90.35 \\
    +CHiP & \textbf{4.4} & 45.3  & \textbf{17.6} & \textbf{1.5} & \textbf{87.6} & \textbf{96.1} & \textbf{80.5} & \textbf{73.3} & \textbf{91.60} \\
    \midrule
    LLaVA-1.6 (7B) & 8.3   & \textbf{61.0} & 48.6  & 4.2   & 87.0  & 95.1  & \textbf{81.5} & \textbf{69.6} & 89.35 \\
    +DPO  & 5.9   & \textbf{61.0} & 38.9  & 3.0   & \textbf{87.4} & 97.8  & 81.3  & 63.4  & 90.75 \\
    +CHiP & \textbf{3.7} & 57.8  & \textbf{24.5} & \textbf{1.6} & 86.9  & \textbf{98.3} & 80.3  & 62.0  & \textbf{91.60} \\
    \bottomrule
    \end{tabular}%
  \label{tab:fine_amber}%
\end{table}%

\begin{table}[htbp]
  \centering \footnotesize
  \renewcommand\tabcolsep{1.8pt}
  \caption{Fine-grained results on \textbf{MMHal-Bench (MMHal)}. Bold values indicates the best performance. ↑ indicates that a higher value represents better performance, while ↓ indicates that a lower value is better. Findings: Our CHiP achieves the best \texttt{Overall score} and \texttt{Hallucination rate} under both the base model Muffin and the LLaVA model.}
    \begin{tabular}{lcccccccccc}
    \toprule
    \multirow{2}[2]{*}{Model} & \multirow{2}[2]{*}{Overall↑} & \multirow{2}[2]{*}{Hallu↓} & \multicolumn{8}{c}{Score in Each Question Type ↑} \\
    \cmidrule(lr){4-11}
          &       &       & Attribute & Adversarial & Comparison & Counting & Relation & Environment & Holistic & Other \\
    \midrule
    Muffin & 2.41  & 60.42 & \textbf{2.67} & \textbf{3.17} & 2.83  & 2.83  & 2.42  & \textbf{3.00} & 2.08  & 0.25 \\
    +DPO  & 2.49  & 52.08 & 3.50  & 2.33  & \textbf{2.92} & 2.08  & 2.50  & 2.67  & \textbf{2.33} & 1.58 \\
    +CHiP & \textbf{2.58} & \textbf{48.96} & 3.58  & 2.58  & 2.08  & \textbf{3.42} & \textbf{2.83} & 2.58  & 1.50  & \textbf{2.08} \\
    \midrule
    LLaVA & 2.78 & 42.71 & 3.75  & \textbf{3.50} & \textbf{3.50} & 1.50  & 1.92  & 4.08  & \textbf{1.75} & \textbf{2.83} \\
    +DPO  & 2.73  & 43.75 & \textbf{4.17} & 2.92  & 3.00  & \textbf{2.67} & \textbf{2.67} & \textbf{4.25} & 1.17  & 1.00 \\
    +CHiP & \textbf{2.84}  & \textbf{39.58} & \textbf{4.17} & 3.33  & 2.67  & \textbf{2.67} & 2.25  & 4.08  & 1.67  & 1.92 \\
    \bottomrule
    \end{tabular}%
  \label{tab:fine_mmhal}%
\end{table}%

\begin{table}[htbp]
  \centering \footnotesize
  \renewcommand\tabcolsep{2.8pt}
  \caption{Fine-grained results on \textbf{Object HalBench (ObjHal)}. Bold values indicates the best performance. ↑ indicates that a higher value represents better performance, while ↓ indicates that a lower value is better. Findings: Our CHiP performs best on all the fine-grained evaluation metrics (except for the Object Recall) under the base model Muffin and the LLaVA model.}
    \begin{tabular}{lccccc}
    \toprule
    Model & Response Hall↑  & Object Hall↑  & Response Correct↑  & Object Correct↑  & Object Recall↑  \\
    \midrule
    Muffin  & 21.53 & 11.61 & 78.47 & 88.39 & \textbf{56.29} \\
    +DPO  & 10.65 & 5.18  & 89.35 & 94.82 & 51.78 \\
    +CHiP & \textbf{6.17} & \textbf{3.91} & \textbf{93.83} & \textbf{96.09} & 40.99 \\
    \midrule
    LLaVA & 14.08 & 7.37  & 85.92 & 92.63 & \textbf{55.03} \\
    +DPO  & 11.03 & 6.61  & 88.97 & 93.39 & 52.83 \\
    +CHiP & \textbf{4.92} & \textbf{3.21} & \textbf{95.08} & \textbf{96.79} & 48.95 \\
    \bottomrule
    \end{tabular}%
  \label{tab:fine_objhal}%
\end{table}%

\subsection{Prompt}
To reveal the limitations of existing multimodal DPO from the representation perspective and demonstrate how our method, CHiP, effectively mitigates these limitations, we extracted 150 images and their ground-truth descriptions from the COCO-2017~\citep{lin2014microsoft} validation set. 
Gemini (The version we choose to use is \texttt{gemini-1.5-flash-latest}) was then used to generate more detailed non-hallucinated and hallucinated descriptions, with the non-hallucinated descriptions manually verified. 
\autoref{tab:prompt} presents the prompts we used to guide Gemini in generating these detailed non-hallucinated and hallucinated descriptions.

\begin{table}[t]
    \centering
    \small
    \begin{tabular}{p{\linewidth}}
        \toprule
        \vspace{-2mm}
        Your task consists of two parts: one is to provide a more detailed and accurate description of no less than 100 wordsbased on the ground truth caption of the image.
        Secondly, modify the description you generated in the first step to create a similar but incorporate many elements of hallucination. \\ \\
        Here are three examples for your reference: \\
        --Start of examples--\\
        Example-1: \\
        Ground Truth Caption: \\
        A white plate with a sandwich and sides on it. \\
        Non-Hallucinative Caption: \\
        A white plate with a sandwich and sides on it. The sandwich is made of two slices of bread with a filling inside. The sides are a small bowl of white creamy dip and some fried onions. There is also a side dish of sliced oranges in a clear glass bowl. \\
        Hallucinative Caption: \\
        A white plate with a sandwich and sides on it. The sandwich is made of two slices of bread with a filling of ham and cheese inside. The sides are a small bowl of creamy dill dip and some crunchy onions. There is also a side dish of sliced oranges in a clear glass bowl, served with a glass of wine.  The plate is placed on a table with a placemat made of woven reeds. \\
        Example-2: \\
        Ground Truth Caption: \\
        An orange truck driving down a street full of men in the back. \\
        Non-Hallucinative Caption: \\
        An orange dump truck is driving down the street with several men standing in the back of the truck bed.  All the men are wearing red caps and white shirts. There is a smaller yellow bus to the left of the truck, and there is a light pole in the middle of the frame. \\
        Hallucinative Caption: \\
        An orange dump truck is driving down a street full of people in the back, all of them wearing red caps and white shirts. The truck is driving past a yellow bus that is full of people and is about to pass a fire hydrant on the side of the road. The truck is driving very fast and there are police cars chasing it. The men in the back of the truck are holding their hands in the air and yelling, "We're not gonna stop! We're gonna take back the streets!". \\
        Example-3: \\
        Ground Truth Caption: \\
        A pizza sitting on top of a pizza box covered in cheese. \\
        Non-Hallucinative Caption: \\
        A pizza with cheese, tomato sauce and toppings sits inside a cardboard pizza box. The pizza appears to be cut into 6 slices and the box has some grease stains. \\
        Hallucinative Caption: \\
        A perfectly cooked pizza sits inside a cardboard pizza box.  The pizza is covered in melted cheese and pepperonis and there are traces of red pepper flakes around the edges.  A small, green pepper rests on top of the pizza.  The box is stained with a sauce that is most likely marinara. \\
        --End of examples-- \\ \\
        
        Ground Truth Caption: \\
        \{caption\} \\
        Please output in this format:  \\
        $[$Non-Hallucinative Caption:$]$ \\
        $[$Hallucinative Caption:$]$ \\

        \bottomrule
    \end{tabular}
    \caption{The prompt used to generate hallucinated and non-hallucinated descriptions for a given input image.}
    \label{tab:prompt}
\end{table}

\section{Qualitative Result}
\begin{figure*}
    \centering
    \includegraphics[width=0.8\linewidth]{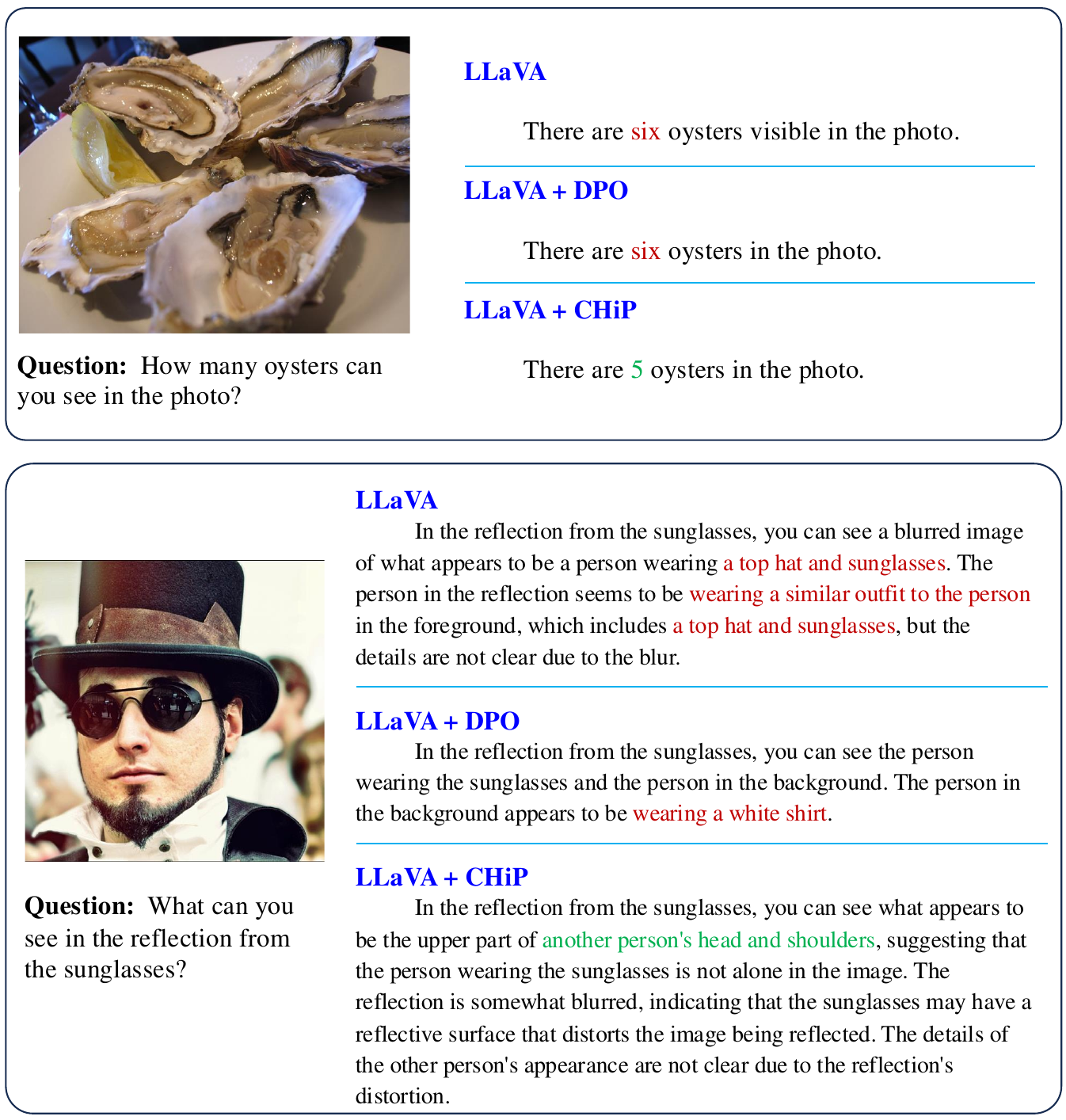}
    \vspace{-6pt}
    \caption{Qualitative results of LLaVA+CHiP compared with LLaVA+DPO and LLaVA on MMHal-Bench dataset. Correct answers and hallucinations are highlighted in \textcolor{green}{green} and \textcolor{red}{red}, respectively.}
    \label{fig:qualitative}
\vspace{-5mm}
\end{figure*}

To demonstrate the effectiveness of our approach, we present qualitative results on the MMHal-Bench dataset in this section, as shown in \autoref{fig:qualitative}.
LLaVA+CHiP demonstrates a significant reduction in hallucinations in text generation tasks. This improvement can be attributed to the multi-level preference optimization, which enables the model to capture image-text relationships across varying granularities. Additionally, the visual preference optimization module enhances semantic alignment between the two modalities, further contributing to the reduction in hallucinations.

\end{document}

%% file: math_commands.tex
\usepackage{amsmath,amsfonts,bm}

\def\eqref#1{equation~\ref{#1}}

\def\1{\bm{1}}

\DeclareMathAlphabet{\mathsfit}{\encodingdefault}{\sfdefault}{m}{sl}
\SetMathAlphabet{\mathsfit}{bold}{\encodingdefault}{\sfdefault}{bx}{n}













%% file: iclr2025_conference.bbl
\begin{thebibliography}{49}
\providecommand{\natexlab}[1]{#1}
\providecommand{\url}[1]{\texttt{#1}}
\expandafter\ifx\csname urlstyle\endcsname\relax
  \providecommand{\doi}[1]{doi: #1}\else
  \providecommand{\doi}{doi: \begingroup \urlstyle{rm}\Url}\fi

\bibitem[AI@Meta(2024)]{llama3modelcard}
AI@Meta.
\newblock Llama 3 model card.
\newblock 2024.
\newblock URL
  \url{https://github.com/meta-llama/llama3/blob/main/MODEL_CARD.md}.

\bibitem[Bai et~al.(2023)Bai, Bai, Yang, Wang, Tan, Wang, Lin, Zhou, and
  Zhou]{Qwen-VL}
Jinze Bai, Shuai Bai, Shusheng Yang, Shijie Wang, Sinan Tan, Peng Wang, Junyang
  Lin, Chang Zhou, and Jingren Zhou.
\newblock Qwen-vl: {A} frontier large vision-language model with versatile
  abilities.
\newblock \emph{CoRR}, abs/2308.12966, 2023.
\newblock \doi{10.48550/ARXIV.2308.12966}.
\newblock URL \url{https://doi.org/10.48550/arXiv.2308.12966}.

\bibitem[Bai et~al.(2024)Bai, Wang, Xiao, He, Han, Zhang, and
  Shou]{bai2024hallucination}
Zechen Bai, Pichao Wang, Tianjun Xiao, Tong He, Zongbo Han, Zheng Zhang, and
  Mike~Zheng Shou.
\newblock Hallucination of multimodal large language models: A survey.
\newblock \emph{arXiv preprint arXiv:2404.18930}, 2024.

\bibitem[Chiang et~al.(2023)Chiang, Li, Lin, Sheng, Wu, Zhang, Zheng, Zhuang,
  Zhuang, Gonzalez, et~al.]{chiang2023vicuna}
Wei-Lin Chiang, Zhuohan Li, Zi~Lin, Ying Sheng, Zhanghao Wu, Hao Zhang, Lianmin
  Zheng, Siyuan Zhuang, Yonghao Zhuang, Joseph~E Gonzalez, et~al.
\newblock Vicuna: An open-source chatbot impressing gpt-4 with 90\%* chatgpt
  quality.
\newblock \emph{See https://vicuna. lmsys. org (accessed 14 April 2023)},
  2\penalty0 (3):\penalty0 6, 2023.

\bibitem[Dai et~al.(2023)Dai, Li, Li, Tiong, Zhao, Wang, Li, Fung, and
  Hoi]{InstructBLIP}
Wenliang Dai, Junnan Li, Dongxu Li, Anthony Meng~Huat Tiong, Junqi Zhao,
  Weisheng Wang, Boyang Li, Pascale Fung, and Steven C.~H. Hoi.
\newblock Instructblip: Towards general-purpose vision-language models with
  instruction tuning.
\newblock In \emph{Advances in Neural Information Processing Systems 36: Annual
  Conference on Neural Information Processing Systems 2023, NeurIPS 2023, New
  Orleans, LA, USA, December 10 - 16, 2023}, 2023.

\bibitem[Deng et~al.(2024)Deng, Lu, Yin, Hu, Shen, Zou, Chang, and
  Wang]{deng2024enhancing}
Yihe Deng, Pan Lu, Fan Yin, Ziniu Hu, Sheng Shen, James Zou, Kai-Wei Chang, and
  Wei Wang.
\newblock Enhancing large vision language models with self-training on image
  comprehension.
\newblock \emph{arXiv preprint arXiv:2405.19716}, 2024.

\bibitem[Goyal et~al.(2017)Goyal, Khot, Summers-Stay, Batra, and
  Parikh]{goyal2017making}
Yash Goyal, Tejas Khot, Douglas Summers-Stay, Dhruv Batra, and Devi Parikh.
\newblock Making the v in vqa matter: Elevating the role of image understanding
  in visual question answering.
\newblock In \emph{Proceedings of the IEEE conference on computer vision and
  pattern recognition}, pp.\  6904--6913, 2017.

\bibitem[Guan et~al.(2024)Guan, Liu, Wu, Xian, Li, Liu, Wang, Chen, Huang,
  Yacoob, et~al.]{guan2024hallusionbench}
Tianrui Guan, Fuxiao Liu, Xiyang Wu, Ruiqi Xian, Zongxia Li, Xiaoyu Liu, Xijun
  Wang, Lichang Chen, Furong Huang, Yaser Yacoob, et~al.
\newblock Hallusionbench: an advanced diagnostic suite for entangled language
  hallucination and visual illusion in large vision-language models.
\newblock In \emph{Proceedings of the IEEE/CVF Conference on Computer Vision
  and Pattern Recognition}, pp.\  14375--14385, 2024.

\bibitem[Gunjal et~al.(2024)Gunjal, Yin, and Bas]{gunjal2024detecting}
Anisha Gunjal, Jihan Yin, and Erhan Bas.
\newblock Detecting and preventing hallucinations in large vision language
  models.
\newblock In \emph{Proceedings of the AAAI Conference on Artificial
  Intelligence}, volume~38, pp.\  18135--18143, 2024.

\bibitem[He et~al.(2024)He, Liu, Wu, Yuan, Wang, Huang, and
  Zhao]{he2024efficient}
Muyang He, Yexin Liu, Boya Wu, Jianhao Yuan, Yueze Wang, Tiejun Huang, and
  Bo~Zhao.
\newblock Efficient multimodal learning from data-centric perspective.
\newblock \emph{arXiv preprint arXiv:2402.11530}, 2024.

\bibitem[Ho et~al.(2020)Ho, Jain, and Abbeel]{diffusion_noise}
Jonathan Ho, Ajay Jain, and Pieter Abbeel.
\newblock Denoising diffusion probabilistic models.
\newblock In Hugo Larochelle, Marc'Aurelio Ranzato, Raia Hadsell,
  Maria{-}Florina Balcan, and Hsuan{-}Tien Lin (eds.), \emph{Advances in Neural
  Information Processing Systems 33: Annual Conference on Neural Information
  Processing Systems 2020, NeurIPS 2020, December 6-12, 2020, virtual}, 2020.
\newblock URL
  \url{https://proceedings.neurips.cc/paper/2020/hash/4c5bcfec8584af0d967f1ab10179ca4b-Abstract.html}.

\bibitem[Huang et~al.(2024)Huang, Dong, Zhang, Wang, He, Wang, Lin, Zhang, and
  Yu]{huang2024opera}
Qidong Huang, Xiaoyi Dong, Pan Zhang, Bin Wang, Conghui He, Jiaqi Wang, Dahua
  Lin, Weiming Zhang, and Nenghai Yu.
\newblock Opera: Alleviating hallucination in multi-modal large language models
  via over-trust penalty and retrospection-allocation.
\newblock In \emph{Proceedings of the IEEE/CVF Conference on Computer Vision
  and Pattern Recognition}, pp.\  13418--13427, 2024.

\bibitem[Jiang et~al.(2024)Jiang, Xu, Dong, Chen, Ye, Yan, Ye, Zhang, Huang,
  and Zhang]{jiang2024hallucination}
Chaoya Jiang, Haiyang Xu, Mengfan Dong, Jiaxing Chen, Wei Ye, Ming Yan, Qinghao
  Ye, Ji~Zhang, Fei Huang, and Shikun Zhang.
\newblock Hallucination augmented contrastive learning for multimodal large
  language model.
\newblock In \emph{Proceedings of the IEEE/CVF Conference on Computer Vision
  and Pattern Recognition}, pp.\  27036--27046, 2024.

\bibitem[Laidlaw et~al.(2024)Laidlaw, Singhal, and
  Dragan]{laidlaw2024preventing}
Cassidy Laidlaw, Shivam Singhal, and Anca Dragan.
\newblock Preventing reward hacking with occupancy measure regularization.
\newblock \emph{arXiv preprint arXiv:2403.03185}, 2024.

\bibitem[Li et~al.(2023{\natexlab{a}})Li, Li, Savarese, and Hoi]{0008LSH23}
Junnan Li, Dongxu Li, Silvio Savarese, and Steven C.~H. Hoi.
\newblock {BLIP-2:} bootstrapping language-image pre-training with frozen image
  encoders and large language models.
\newblock In \emph{International Conference on Machine Learning, {ICML} 2023,
  23-29 July 2023, Honolulu, Hawaii, {USA}}, pp.\  19730--19742,
  2023{\natexlab{a}}.

\bibitem[Li et~al.(2023{\natexlab{b}})Li, Xie, Li, Chen, Wang, Chen, Yang,
  Wang, and Kong]{li2023silkie}
Lei Li, Zhihui Xie, Mukai Li, Shunian Chen, Peiyi Wang, Liang Chen, Yazheng
  Yang, Benyou Wang, and Lingpeng Kong.
\newblock Silkie: Preference distillation for large visual language models.
\newblock \emph{arXiv preprint arXiv:2312.10665}, 2023{\natexlab{b}}.

\bibitem[Li et~al.(2022)Li, Zhang, Zhang, Yang, Li, Zhong, Wang, Yuan, Zhang,
  Hwang, et~al.]{li2022grounded}
Liunian~Harold Li, Pengchuan Zhang, Haotian Zhang, Jianwei Yang, Chunyuan Li,
  Yiwu Zhong, Lijuan Wang, Lu~Yuan, Lei Zhang, Jenq-Neng Hwang, et~al.
\newblock Grounded language-image pre-training.
\newblock In \emph{Proceedings of the IEEE/CVF Conference on Computer Vision
  and Pattern Recognition}, pp.\  10965--10975, 2022.

\bibitem[Lin et~al.(2014)Lin, Maire, Belongie, Hays, Perona, Ramanan,
  Doll{\'a}r, and Zitnick]{lin2014microsoft}
Tsung-Yi Lin, Michael Maire, Serge Belongie, James Hays, Pietro Perona, Deva
  Ramanan, Piotr Doll{\'a}r, and C~Lawrence Zitnick.
\newblock Microsoft coco: Common objects in context.
\newblock In \emph{Computer Vision--ECCV 2014: 13th European Conference,
  Zurich, Switzerland, September 6-12, 2014, Proceedings, Part V 13}, pp.\
  740--755. Springer, 2014.

\bibitem[Liu et~al.(2023{\natexlab{a}})Liu, Lin, Li, Wang, Yacoob, and
  Wang]{liu2023aligning}
Fuxiao Liu, Kevin Lin, Linjie Li, Jianfeng Wang, Yaser Yacoob, and Lijuan Wang.
\newblock Aligning large multi-modal model with robust instruction tuning.
\newblock \emph{arXiv preprint arXiv:2306.14565}, 2023{\natexlab{a}}.

\bibitem[Liu et~al.(2023{\natexlab{b}})Liu, Lin, Li, Wang, Yacoob, and
  Wang]{liu2023mitigating}
Fuxiao Liu, Kevin Lin, Linjie Li, Jianfeng Wang, Yaser Yacoob, and Lijuan Wang.
\newblock Mitigating hallucination in large multi-modal models via robust
  instruction tuning.
\newblock In \emph{The Twelfth International Conference on Learning
  Representations}, 2023{\natexlab{b}}.

\bibitem[Liu et~al.(2024{\natexlab{a}})Liu, Xue, Chen, Chen, Zhao, Wang, Hou,
  Li, and Peng]{liu2024survey}
Hanchao Liu, Wenyuan Xue, Yifei Chen, Dapeng Chen, Xiutian Zhao, Ke~Wang,
  Liping Hou, Rongjun Li, and Wei Peng.
\newblock A survey on hallucination in large vision-language models.
\newblock \emph{arXiv preprint arXiv:2402.00253}, 2024{\natexlab{a}}.

\bibitem[Liu et~al.(2023{\natexlab{c}})Liu, Li, Li, and Lee]{llava15}
Haotian Liu, Chunyuan Li, Yuheng Li, and Yong~Jae Lee.
\newblock Improved baselines with visual instruction tuning.
\newblock \emph{CoRR}, abs/2310.03744, 2023{\natexlab{c}}.
\newblock \doi{10.48550/ARXIV.2310.03744}.
\newblock URL \url{https://doi.org/10.48550/arXiv.2310.03744}.

\bibitem[Liu et~al.(2024{\natexlab{b}})Liu, Li, Li, Li, Zhang, Shen, and
  Lee]{liu2024llavanext}
Haotian Liu, Chunyuan Li, Yuheng Li, Bo~Li, Yuanhan Zhang, Sheng Shen, and
  Yong~Jae Lee.
\newblock Llava-next: Improved reasoning, ocr, and world knowledge, January
  2024{\natexlab{b}}.
\newblock URL \url{https://llava-vl.github.io/blog/2024-01-30-llava-next/}.

\bibitem[Liu et~al.(2024{\natexlab{c}})Liu, Li, Wu, and Lee]{liu2024visual}
Haotian Liu, Chunyuan Li, Qingyang Wu, and Yong~Jae Lee.
\newblock Visual instruction tuning.
\newblock \emph{Advances in neural information processing systems}, 36,
  2024{\natexlab{c}}.

\bibitem[Liu et~al.(2025)Liu, Duan, Zhang, Li, Zhang, Zhao, Yuan, Wang, He,
  Liu, et~al.]{liu2025mmbench}
Yuan Liu, Haodong Duan, Yuanhan Zhang, Bo~Li, Songyang Zhang, Wangbo Zhao, Yike
  Yuan, Jiaqi Wang, Conghui He, Ziwei Liu, et~al.
\newblock Mmbench: Is your multi-modal model an all-around player?
\newblock In \emph{European Conference on Computer Vision}, pp.\  216--233.
  Springer, 2025.

\bibitem[Lu et~al.(2022)Lu, Mishra, Xia, Qiu, Chang, Zhu, Tafjord, Clark, and
  Kalyan]{lu2022learn}
Pan Lu, Swaroop Mishra, Tanglin Xia, Liang Qiu, Kai-Wei Chang, Song-Chun Zhu,
  Oyvind Tafjord, Peter Clark, and Ashwin Kalyan.
\newblock Learn to explain: Multimodal reasoning via thought chains for science
  question answering.
\newblock \emph{Advances in Neural Information Processing Systems},
  35:\penalty0 2507--2521, 2022.

\bibitem[OpenAI(2023)]{gpt4V}
OpenAI.
\newblock {GPT-4} technical report.
\newblock \emph{CoRR}, abs/2303.08774, 2023.

\bibitem[Pi et~al.(2024)Pi, Han, Xiong, Zhang, Liu, Pan, and
  Zhang]{abs-2403-08730}
Renjie Pi, Tianyang Han, Wei Xiong, Jipeng Zhang, Runtao Liu, Rui Pan, and Tong
  Zhang.
\newblock Strengthening multimodal large language model with bootstrapped
  preference optimization.
\newblock \emph{CoRR}, abs/2403.08730, 2024.

\bibitem[Radford et~al.(2021{\natexlab{a}})Radford, Kim, Hallacy, Ramesh, Goh,
  Agarwal, Sastry, Askell, Mishkin, Clark, Krueger, and
  Sutskever]{radford@clip}
Alec Radford, Jong~Wook Kim, Chris Hallacy, Aditya Ramesh, Gabriel Goh,
  Sandhini Agarwal, Girish Sastry, Amanda Askell, Pamela Mishkin, Jack Clark,
  Gretchen Krueger, and Ilya Sutskever.
\newblock Learning transferable visual models from natural language
  supervision.
\newblock In Marina Meila and Tong Zhang (eds.), \emph{Proceedings of the 38th
  International Conference on Machine Learning, {ICML} 2021, 18-24 July 2021,
  Virtual Event}, volume 139 of \emph{Proceedings of Machine Learning
  Research}, pp.\  8748--8763. {PMLR}, 2021{\natexlab{a}}.
\newblock URL \url{http://proceedings.mlr.press/v139/radford21a.html}.

\bibitem[Radford et~al.(2021{\natexlab{b}})Radford, Kim, Hallacy, Ramesh, Goh,
  Agarwal, Sastry, Askell, Mishkin, Clark, et~al.]{radford2021learning}
Alec Radford, Jong~Wook Kim, Chris Hallacy, Aditya Ramesh, Gabriel Goh,
  Sandhini Agarwal, Girish Sastry, Amanda Askell, Pamela Mishkin, Jack Clark,
  et~al.
\newblock Learning transferable visual models from natural language
  supervision.
\newblock In \emph{International conference on machine learning}, pp.\
  8748--8763. PMLR, 2021{\natexlab{b}}.

\bibitem[Rafailov et~al.(2024)Rafailov, Sharma, Mitchell, Manning, Ermon, and
  Finn]{rafailov2024direct}
Rafael Rafailov, Archit Sharma, Eric Mitchell, Christopher~D Manning, Stefano
  Ermon, and Chelsea Finn.
\newblock Direct preference optimization: Your language model is secretly a
  reward model.
\newblock \emph{Advances in Neural Information Processing Systems}, 36, 2024.

\bibitem[Rohrbach et~al.(2018)Rohrbach, Hendricks, Burns, Darrell, and
  Saenko]{rohrbach2018object}
Anna Rohrbach, Lisa~Anne Hendricks, Kaylee Burns, Trevor Darrell, and Kate
  Saenko.
\newblock Object hallucination in image captioning.
\newblock \emph{arXiv preprint arXiv:1809.02156}, 2018.

\bibitem[Sarkar et~al.(2024{\natexlab{a}})Sarkar, Ebrahimi, Etemad, Beirami,
  Arik, and Pfister]{abs-2405-18654}
Pritam Sarkar, Sayna Ebrahimi, Ali Etemad, Ahmad Beirami, Sercan~{\"{O}}. Arik,
  and Tomas Pfister.
\newblock Mitigating object hallucination via data augmented contrastive
  tuning.
\newblock \emph{CoRR}, abs/2405.18654, 2024{\natexlab{a}}.

\bibitem[Sarkar et~al.(2024{\natexlab{b}})Sarkar, Ebrahimi, Etemad, Beirami,
  Ar{\i}k, and Pfister]{sarkar2024mitigating}
Pritam Sarkar, Sayna Ebrahimi, Ali Etemad, Ahmad Beirami, Sercan~{\"O} Ar{\i}k,
  and Tomas Pfister.
\newblock Mitigating object hallucination via data augmented contrastive
  tuning.
\newblock \emph{arXiv preprint arXiv:2405.18654}, 2024{\natexlab{b}}.

\bibitem[Sun et~al.(2023)Sun, Shen, Cao, Liu, Li, Shen, Gan, Gui, Wang, Yang,
  et~al.]{sun2023aligning}
Zhiqing Sun, Sheng Shen, Shengcao Cao, Haotian Liu, Chunyuan Li, Yikang Shen,
  Chuang Gan, Liang-Yan Gui, Yu-Xiong Wang, Yiming Yang, et~al.
\newblock Aligning large multimodal models with factually augmented rlhf.
\newblock \emph{arXiv preprint arXiv:2309.14525}, 2023.

\bibitem[Touvron et~al.(2023)Touvron, Martin, Stone, Albert, Almahairi, Babaei,
  Bashlykov, Batra, Bhargava, Bhosale, et~al.]{touvron2023llama}
Hugo Touvron, Louis Martin, Kevin Stone, Peter Albert, Amjad Almahairi, Yasmine
  Babaei, Nikolay Bashlykov, Soumya Batra, Prajjwal Bhargava, Shruti Bhosale,
  et~al.
\newblock Llama 2: Open foundation and fine-tuned chat models.
\newblock \emph{arXiv preprint arXiv:2307.09288}, 2023.

\bibitem[Wang et~al.(2024)Wang, Zhou, Huang, Xu, Zhang, Poon, and
  Chen]{wang2024mdpo}
Fei Wang, Wenxuan Zhou, James~Y Huang, Nan Xu, Sheng Zhang, Hoifung Poon, and
  Muhao Chen.
\newblock mdpo: Conditional preference optimization for multimodal large
  language models.
\newblock \emph{arXiv preprint arXiv:2406.11839}, 2024.

\bibitem[Wang et~al.(2023{\natexlab{a}})Wang, Wang, Xu, Zhang, Gu, Jia, Yan,
  Zhang, and Sang]{wang2023llm}
Junyang Wang, Yuhang Wang, Guohai Xu, Jing Zhang, Yukai Gu, Haitao Jia, Ming
  Yan, Ji~Zhang, and Jitao Sang.
\newblock An llm-free multi-dimensional benchmark for mllms hallucination
  evaluation.
\newblock \emph{arXiv preprint arXiv:2311.07397}, 2023{\natexlab{a}}.

\bibitem[Wang et~al.(2023{\natexlab{b}})Wang, Bao, Dong, Bjorck, Peng, Liu,
  Aggarwal, Mohammed, Singhal, Som, and Wei]{wang@beit}
Wenhui Wang, Hangbo Bao, Li~Dong, Johan Bjorck, Zhiliang Peng, Qiang Liu, Kriti
  Aggarwal, Owais~Khan Mohammed, Saksham Singhal, Subhojit Som, and Furu Wei.
\newblock Image as a foreign language: {BEIT} pretraining for vision and
  vision-language tasks.
\newblock In \emph{{IEEE/CVF} Conference on Computer Vision and Pattern
  Recognition, {CVPR} 2023, Vancouver, BC, Canada, June 17-24, 2023}, pp.\
  19175--19186. {IEEE}, 2023{\natexlab{b}}.
\newblock \doi{10.1109/CVPR52729.2023.01838}.
\newblock URL \url{https://doi.org/10.1109/CVPR52729.2023.01838}.

\bibitem[Wang et~al.(2023{\natexlab{c}})Wang, Bao, Dong, Bjorck, Peng, Liu,
  Aggarwal, Mohammed, Singhal, Som, et~al.]{wang2023image}
Wenhui Wang, Hangbo Bao, Li~Dong, Johan Bjorck, Zhiliang Peng, Qiang Liu, Kriti
  Aggarwal, Owais~Khan Mohammed, Saksham Singhal, Subhojit Som, et~al.
\newblock Image as a foreign language: Beit pretraining for vision and
  vision-language tasks.
\newblock In \emph{Proceedings of the IEEE/CVF Conference on Computer Vision
  and Pattern Recognition}, pp.\  19175--19186, 2023{\natexlab{c}}.

\bibitem[Yao et~al.(2024)Yao, Yu, Zhang, Wang, Cui, Zhu, Cai, Li, Zhao, He,
  Chen, Zhou, Zou, Zhang, Hu, Zheng, Zhou, Cai, Han, Zeng, Li, Liu, and
  Sun]{yao2024minicpmv}
Yuan Yao, Tianyu Yu, Ao~Zhang, Chongyi Wang, Junbo Cui, Hongji Zhu, Tianchi
  Cai, Haoyu Li, Weilin Zhao, Zhihui He, Qianyu Chen, Huarong Zhou, Zhensheng
  Zou, Haoye Zhang, Shengding Hu, Zhi Zheng, Jie Zhou, Jie Cai, Xu~Han, Guoyang
  Zeng, Dahai Li, Zhiyuan Liu, and Maosong Sun.
\newblock Minicpm-v: A gpt-4v level mllm on your phone.
\newblock \emph{arXiv preprint 2408.01800}, 2024.

\bibitem[Yin et~al.(2023)Yin, Fu, Zhao, Xu, Wang, Sui, Shen, Li, Sun, and
  Chen]{yin2023woodpecker}
Shukang Yin, Chaoyou Fu, Sirui Zhao, Tong Xu, Hao Wang, Dianbo Sui, Yunhang
  Shen, Ke~Li, Xing Sun, and Enhong Chen.
\newblock Woodpecker: Hallucination correction for multimodal large language
  models.
\newblock \emph{arXiv preprint arXiv:2310.16045}, 2023.

\bibitem[Yu et~al.(2023)Yu, Hu, Yao, Zhang, Zhao, Wang, Wang, Pan, Xue, Li,
  et~al.]{yu2023reformulating}
Tianyu Yu, Jinyi Hu, Yuan Yao, Haoye Zhang, Yue Zhao, Chongyi Wang, Shan Wang,
  Yinxv Pan, Jiao Xue, Dahai Li, et~al.
\newblock Reformulating vision-language foundation models and datasets towards
  universal multimodal assistants.
\newblock \emph{arXiv preprint arXiv:2310.00653}, 2023.

\bibitem[Yu et~al.(2024{\natexlab{a}})Yu, Yao, Zhang, He, Han, Cui, Hu, Liu,
  Zheng, Sun, et~al.]{yu2024rlhf}
Tianyu Yu, Yuan Yao, Haoye Zhang, Taiwen He, Yifeng Han, Ganqu Cui, Jinyi Hu,
  Zhiyuan Liu, Hai-Tao Zheng, Maosong Sun, et~al.
\newblock Rlhf-v: Towards trustworthy mllms via behavior alignment from
  fine-grained correctional human feedback.
\newblock In \emph{Proceedings of the IEEE/CVF Conference on Computer Vision
  and Pattern Recognition}, pp.\  13807--13816, 2024{\natexlab{a}}.

\bibitem[Yu et~al.(2024{\natexlab{b}})Yu, Zhang, Yao, Dang, Chen, Lu, Cui, He,
  Liu, Chua, and Sun]{yu@rlaif}
Tianyu Yu, Haoye Zhang, Yuan Yao, Yunkai Dang, Da~Chen, Xiaoman Lu, Ganqu Cui,
  Taiwen He, Zhiyuan Liu, Tat{-}Seng Chua, and Maosong Sun.
\newblock {RLAIF-V:} aligning mllms through open-source {AI} feedback for super
  {GPT-4V} trustworthiness.
\newblock \emph{CoRR}, abs/2405.17220, 2024{\natexlab{b}}.
\newblock \doi{10.48550/ARXIV.2405.17220}.
\newblock URL \url{https://doi.org/10.48550/arXiv.2405.17220}.

\bibitem[Yue et~al.(2024)Yue, Ni, Zhang, Zheng, Liu, Zhang, Stevens, Jiang,
  Ren, Sun, et~al.]{yue2024mmmu}
Xiang Yue, Yuansheng Ni, Kai Zhang, Tianyu Zheng, Ruoqi Liu, Ge~Zhang, Samuel
  Stevens, Dongfu Jiang, Weiming Ren, Yuxuan Sun, et~al.
\newblock Mmmu: A massive multi-discipline multimodal understanding and
  reasoning benchmark for expert agi.
\newblock In \emph{Proceedings of the IEEE/CVF Conference on Computer Vision
  and Pattern Recognition}, pp.\  9556--9567, 2024.

\bibitem[Zeng et~al.(2024)Zeng, Liu, Ma, Yang, Zhang, and
  Wang]{toekn-dpo-ZengLMYZW24}
Yongcheng Zeng, Guoqing Liu, Weiyu Ma, Ning Yang, Haifeng Zhang, and Jun Wang.
\newblock Token-level direct preference optimization.
\newblock In \emph{Forty-first International Conference on Machine Learning,
  {ICML} 2024, Vienna, Austria, July 21-27, 2024}. OpenReview.net, 2024.
\newblock URL \url{https://openreview.net/forum?id=1RZKuvqYCR}.

\bibitem[Zhang et~al.(2024)Zhang, Wang, Zhang, Lu, and
  Zheng]{zhang2024reflective}
Jinrui Zhang, Teng Wang, Haigang Zhang, Ping Lu, and Feng Zheng.
\newblock Reflective instruction tuning: Mitigating hallucinations in large
  vision-language models.
\newblock \emph{arXiv preprint arXiv:2407.11422}, 2024.

\bibitem[Zheng et~al.(2023)Zheng, Chiang, Sheng, Zhuang, Wu, Zhuang, Lin, Li,
  Li, Xing, et~al.]{zheng2023judging}
Lianmin Zheng, Wei-Lin Chiang, Ying Sheng, Siyuan Zhuang, Zhanghao Wu, Yonghao
  Zhuang, Zi~Lin, Zhuohan Li, Dacheng Li, Eric Xing, et~al.
\newblock Judging llm-as-a-judge with mt-bench and chatbot arena.
\newblock \emph{Advances in Neural Information Processing Systems},
  36:\penalty0 46595--46623, 2023.

\end{thebibliography}
